\documentclass[12pt]{article}
\usepackage{graphicx}

\usepackage[ruled,vlined]{algorithm2e}

\usepackage{xfrac}
\usepackage{caption}
\usepackage{subcaption}
\usepackage{amsmath}
\usepackage{amssymb}
\usepackage{color}
\usepackage{enumerate}

\usepackage{scicite}

\usepackage{times}

\newenvironment{sciabstract}{%
\begin{quote} \bf}
{\end{quote}}

\title{Game-theoretical trajectory planning enhances social
acceptability for humans}

\author
{Giada Galati $^{1}$, Stefano Primatesta $^{2}$, Sergio Grammatico $^{3}$, Simone Macr\`i $^{4}$,\\ Alessandro Rizzo $^{1,5\ast}$\\
\normalsize{$^{1}$Department of Electronics and Telecomunications, Politecnico di Torino, Italy}\\
\normalsize{$^{2}$Department of Mechanical and Aerospace Engineering, Politecnico di Torino, Italy}\\
\normalsize{$^{3}$ Delft Center for Systems and Control, TU Delft, The Netherlands} \\
\normalsize{$^{4}$ Department of Cell Biology and Neurosciences, Istituto Superiore di Sanità, Rome, Italy} \\
\normalsize{$^{5}$ Institute for Invention, Innovation, and Entrepreneurship,} \\
\normalsize{New York University Tandon School of Engineering, Brooklyn NY, USA} \\\\
\normalsize{$^\ast$ Corresponding author, E-mail:  alessandro.rizzo@polito.it}
}

\date{}

\begin{document} 


\baselineskip24pt


\maketitle

\begin{sciabstract}

Short title: Social acceptability of game-theoretical planners\\

Abstract -- Since humans and robots are increasingly sharing portions of their operational spaces, experimental evidence is needed to ascertain the safety and social acceptability of robots in human-populated environments.  Although several studies have aimed at devising strategies for robot trajectory planning to perform \emph{safe} motion in populated environments, a few efforts have \emph{measured} to what extent a robot trajectory is  \emph{accepted} by humans. Here, we present a navigation system for autonomous robotics that ensures safety and social acceptability of robotic trajectories. We overcome the typical reactive nature of state-of-the-art trajectory planners by leveraging non-cooperative game theory to design a planner that encapsulates human-like features of preservation of a vital space, recognition of groups, sequential and strategized decision making, and smooth obstacle avoidance. Social acceptability is measured through a variation of the Turing test administered in the form of a survey questionnaire to a pool of 691 participants. Comparison terms for our tests are a state-of-the-art navigation algorithm (Enhanced Vector Field Histogram, VFH) and purely human trajectories. While all participants easily recognized the non-human nature of VFH-generated trajectories, the distinction between game-theoretical trajectories and human ones were hardly revealed. These results mark a strong milestone toward the full integration of robots in social environments. 

\end{sciabstract}

\section*{Summary}

Experimental evidence confirms that our game-theoretical planner improves social acceptability of robotic behavior.

\section*{Introduction}

The widespread diffusion of service robots for diverse applications is making autonomous robots more and more pervasive in our lives~\cite{torras2016service}. In the near future, autonomous robots will likely coexist and share our very space. Application scenarios will be  characterized by populated and dynamic environments, where autonomous navigation has to ensure not only the physical safety of human subjects, but also a great degree of \emph{social acceptability}~\cite{kruse2013human}. Trajectory planners at the state of the art mostly aimed at ensuring the former requisite~\cite{fox1997dynamic,fiorini1998motion,van2008reciprocal}, while seldom tackling the social acceptability issue. Most of contemporary autonomous navigation algorithms model humans as inanimate dynamic obstacles rather than social entities interacting with each other through complex and strategized patterns~\cite{kivrak2020social}. The oversimplification of human behavioral traits in the design of navigation algorithm may have severe consequences, such as the emergence of the well-known  ``freezing robot problem''~\cite{trautman2015robot}. 

Socially-aware navigation is gaining momentum as a fundamental requirement toward the design of \emph{social robots,} able to adhere to social convention toward providing a friendly and comfortable interaction with humans. Socially-aware navigation combines perception, dynamical system theory, social conventions, human motion modeling, and psychology. Trajectories generated in this context should be predictable, adaptable, easily understandable, and acceptable by humans~\cite{rios2015proxemics}. Toward improving trust, comfort, and social acceptance, humans should be explicitly considered by robots as intelligent agents who interact and may influence the motion of others~\cite{turnwald2019human}. Recent efforts in socially-aware navigation model humans as static entities~\cite{sisbot2007human} or as agents driven by very simplistic motion models~\cite{shiomi2014towards}. Such simplistic assumptions may hardly cope with the complexity of human behavior and interaction, yielding trajectories that far from predictable, smooth, and in turn acceptable by humans. Models based on  learning theory, on the other hand, promise better results~\cite{chen2017socially} provided that a large training data set involving human subjects is available, which is not always the case. 

\begin{figure}[h]
        \centering
        \includegraphics[width=0.7 \textwidth] {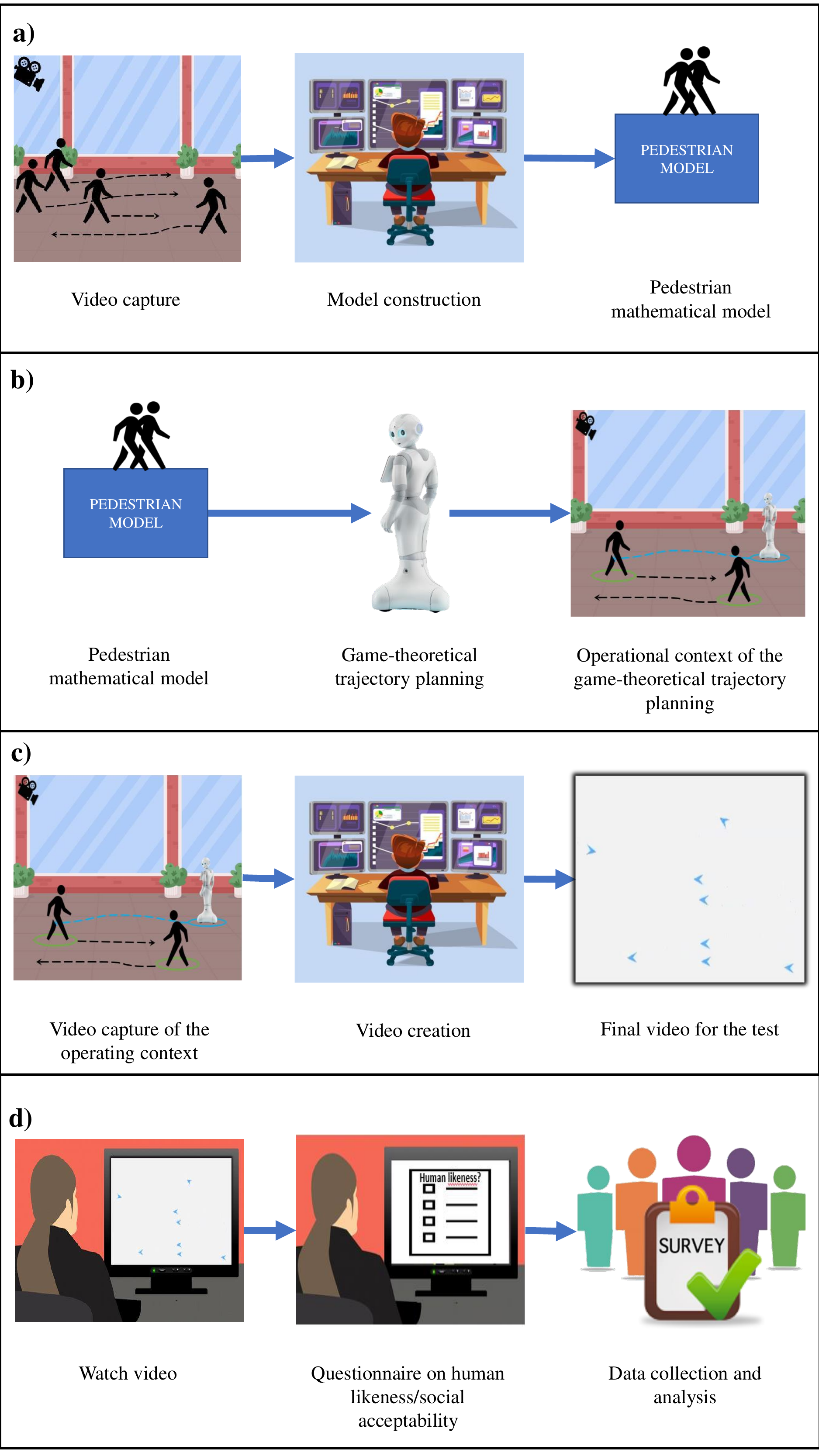}
        \caption[width=1\textwidth]{Graphical abstract of the procedure.  \textbf{a)} Construction of the game-theoretical model for human motion; \textbf{b)} creation of the game-theoretical trajectory planner based on the model previous designed, creation of the virtual  environment  and evaluation of the performance metrics; \textbf{c)} creation of the videos with pedestrians and the robot controlled by our game-theoretical trajectory planner; \textbf{d)} survey questionnaire and data collection.}
         \label{Fig:Intro_Image}
       
    \end{figure}

Here, we present a socially-aware robot navigation strategy that accurately models human behavior using game theory (see Figure~\ref{Fig:Intro_Image} for a graphical abstract of the procedure). Game theory offers substantial benefits compared to alternative modeling methods, such as reactive strategies~\cite{helbing1995social,tadokoro1995motion,hoeller2007accompanying} and learning schemes~\cite{bennewitz2005learning,alahi2016social,gupta2018social,liang2019peeking}. With respect to the former, game theory is able to perform motion prediction and anticipation of the behavior of other humans, typical of human decision making in social contexts~\cite{turnwald2016understanding}. Compared to the latter, it overcomes their distinctive lack of explainability, generalization, and the need for large training data set.  Game theory has successfully found applications in robot motion planning, as in Zhang et al.~\cite{zhang1998motion}, where a non-cooperative, zero-sum game is used to coordinate their motion and avoid obstacles to execute a set of prioritized tasks. Gabler et al.~\cite{gabler2017game} propose a game-theoretical framework in which humans and robots collaborate in an industrial assembly scenarios;  Dragen et al.~\cite{dragan2017robot} and Nikolaidis et al.~\cite{nikolaidis2017mathematical} model the interaction between human and robot as a two-player game and point out how different game assumptions and approximations lead to different robot behaviors.

Our approach uses non-cooperative game theory~\cite{nash1951non} to model the navigation behavior of multiple humans in populated environments, positing that conditions of safe navigation, adherence to social norms, and psychological comfort correspond to a Nash equilibrium in the proposed game-theoretical model. Differently from the previously cited works, our model contemplates more than two players --a feature that is essential to model populated environments. The human motion model informs the design of a robotic trajectory planner, whereby the robot tends to mimic human behavior during motion and interaction in populated environment. Hence, we pursue social acceptability through the concept of anthropomorphism~\cite{epley2007seeing,roesler2021meta} --the intrinsic tendency of humans to attribute intentions and consciousness to non-human entities~\cite{waytz2010sees}. 

Our work marks an important milestone in the field of social robotics. It provides an efficient, social-aware motion planning framework  that encapsulates realistic features of human crowds, remarkably enhancing   the social acceptance of the planned trajectories. Namely, we incorporate the human vital space~\cite{hall1966hidden}, the recognition of human groups~\cite{mavrogiannis2021core}, the sequential decision-making typical of human beings~\cite{xie2017learning}, and a natural human-obstacle interaction~\cite{manual1985special} --features that are often missing in many approaches, including those based on game theory~\cite{turnwald2019human}.

The methodology proposed in this paper is generally applicable to any kind of mobile robot. To avoid confounds related to the choice of specific hardware setup and focus on the assessment of  human perception of the robot motion, validation is executed on virtualized environments, where the humanly populated scene is extrapolated from surveillance videos. Three different experimental conditions are considered: the first involves only human subjects, the second contains a virtualized mobile robot programmed through the state-of-the-art Enhanced Vector Field Histogram (VFH~\cite{ulrich1998vfh+}) algorithm moving through the population, the third replaces the VFH algorithm with our game-theoretical approach. 

Across the three experimental conditions, we perform a twofold validation of our approach: first, we evaluate performance metrics typical of path planning (path length ratio, path regularity, and distance to the closest pedestrian), and then we analyze the results of a survey questionnaire to directly assess social acceptability by human subjects. To this aim, we administered a variant of the Turing test to a pool of 691 volunteers, who evaluated the human likeness of three sets of videos corresponding to the three scenarios explained above. To conceal the appearance of the agents, we masked humans and robots by replacing them with arrows so that the volunteers did not have the possibility to distinguish between them. 

Evidence from our experimental campaign reveals that trajectories generated by our game-theoretical approach exhibit performance metrics that are efficient and closer than those achieved by human subjects than VFH. Moreover, the outcome of the survey questionnaire highlights the superior acceptability of game-theoretical-generated trajectories with respect to those generated through VFH.

\section*{Methods}
 
\subsection*{Game-theoretical model}

 \subparagraph*{Assumptions}
Let us start with a description of all the assumptions supporting our game-theoretical model for human motion. To improve readability, here and henceforth we will refer to human subjects as \emph{agent}. This term will be also used for the robot when no distinction between the two categories is required. 

All pedestrians are \emph{rational} agents with \emph{common knowledge} moving in a 2D \emph{populated dynamic} environment. 

\emph{Rational} behavior entails that agents only aim to reach their own \emph{individual} motion goal. In mathematical terms, this translates into a minimization of an individual cost (equivalently, a maximization of an individual  benefit), such as their overall path length~\cite{bitgood2006not} or energy consumption~\cite{mcneill2002energetics}. Practically, agents continuously update their navigation behavior while walking in populated environments, based on the observation and possible prediction of the motion of the surrounding agents. 

The possession of \emph{common knowledge} by agents in our game-theoretical model implies that all agents have the same knowledge of the characteristics of the environment and of the rules that regulate it. Such an assumption is reasonable when dealing with models of human traits, as  individuals commonly learn these skills by experience during everyday life~\cite{turnwald2019human}. 

Specifically, we consider a  \emph{populated dynamic environment}, possibly busy, but not crowded, such as typical streets occupied by pedestrians walking on sideways, or populated indoor spaces, such as hotel halls~\cite{turnwald2019human}. We suppose that the environment contains static obstacles, that have to be avoided by agents in a \emph{natural} manner. Our approach is based on a microscopic modeling strategy, whereby a single individual is mapped onto a single software agents, which mimics the individuals' decision and their interactions.

    \subparagraph*{Game description}
    \label{description of the game}
    The proposed model for pedestrian motion is a non-cooperative, static, perfect information, finite, and general-sum game with many players (or agents). 
    
    In our model, each agent aims at reaching its own goal \emph{individually}, but the minimization of its individual cost does not exclude the possibility to collaborate with other agents,  should this help attaining  \emph{individual} goals \cite{osborne1994course} as well.
    Cohesive groups of agents are considered as single agents, whereby members of the group share a common strategy and a common motion pattern. This last assumption practically entails that the navigation strategy of the robot in avoiding human groups would treat them as a single entity, without attempting at breaking into them to better attain its own navigation goal.

    The game is \emph{static} in the sense that agents move and take decisions \emph{simultaneously}, it is based on \emph{perfect information}, that is, each agent knows the current and the previous actions of all agents, e.g. via direct observations.

    The game is also \emph{finite}, i.e., the game has a \emph{finite} number $N$ of agents belonging to the agent set $\mathcal{N}$, where each agent ${i \in \mathcal{N}}$ can choose among a \emph{finite} number of actions available, defined with the action set $\Theta$, which is supposed to be common to all agents. In particular, we indicate with $\theta_{i} (t) \in \Theta$ the action executed by agent $i$ at the discrete time $t$. In our application, the execution of action $\theta_i(t)$ corresponds to a motion of agent $i$ in the 2D plane at constant velocity $v$ and constant heading $\theta_i (t)$ over the whole discrete time step $\Delta t$. We assume that agents have a bounded visibility angle and the possible action $\theta_i(t)$ is designed to uniformly partition such an angle. We denote with $p_i \in \mathbb{R}^2$ the position of agent $i$ in the 2D environment, with respect to a fixed orthogonal reference frame.  
    
    Moreover, the proposed model is a \emph{general-sum} game, i.e., the sum of all gains and losses of the utility functions over all agents is \textit{not necessarily} equal to zero.

    Similar to~\cite{turnwald2016understanding}, we postulate that, in such a navigation task, agents tend to reach a Nash equilibrium -- the condition in which no agent has an incentive to unilaterally change its own action (or strategy) if the other agents do not change theirs. In other words, a Nash equilibrium occurs when each agent achieves its best response, i.e., its minimum individual cost, given the actions of the other agents. In general, however, existence and uniqueness of a Nash equilibrium is not guaranteed in our setup, and its analytical characterization is almost always impossible to have, thus making numerical approaches for an approximate computational necessary. Here, the Nash equilibrium is approximately computed via the \emph{sequential best response} approach ~\cite{sagratella2017algorithms}.

    Let us explain the idea of the sequential best response for two agents, A and B:
    agent A observes the motion of agent B and then solves an optimization problem to determine its own strategy, given the latest observed strategy of agent B.
    Afterwards, a check action is performed, verifying if the strategies of both agents are the same as those computed in the previous iteration; in such a case, the game has reached a Nash equilibrium. Otherwise, agent B computes its optimal strategy, given the latest observed strategy of agent A. The procedure is applied iteratively, until the equilibrium condition is met. The same strategy identically extends to $N$ agents. 
    
    Our modeling procedure assumes that all the agents in the planar space play the game mentioned above. After the model has been identified, we will use it to control a single, synthetic agent to navigate through the populated environment. Such an agent is called  \emph{robot player}.
  
    \subparagraph*{Optimization problem}
    \label{Optimization problem}
    
    The sequential best response approach in our game-theoretical model for the human motion in a populated environment requires the solution of a set of interdependent optimization problems, one for each agent moving in the environment. 
    The goal of the optimization problem for each agent \emph{i} is to find the best sequence of actions, $\boldsymbol{\theta^{*}_{i}} = (\theta_{i}(t), \theta_i(t+\Delta t), \theta_i(t+2 \Delta t), \ldots, \theta_i(t+T \Delta t))$, over a finite prediction horizon $T \Delta t$, given the actions of the other agents. Without loss of generalization and to improve readability, here and henceforth we assume a unitary discrete-time step, i.e., $\Delta t = 1$. 
    
    All agents seek for the Nash equilibrium applying the sequential best response strategy, solving their own optimization problem on the basis of the observed behavior of the rest of the population. 
    We define the optimization problem for each agent $i \in \mathcal{N}$ as

       \begin{subequations}
\begin{alignat}{2}
 \boldsymbol{\theta^{*}_{i}} =~ &\!\underset{\boldsymbol{\theta_{i}}}{\mathrm{min}}       &\quad& J(\boldsymbol{\theta_{i}}) \label{eq:optProb}\\
&\text{s.t.} &    &  \left\| p_i(t, \theta_{i}(t))-p_j(t) \right\|_{2}~\geq \beta  \quad \forall ~t, \forall  i ,  j \in \mathcal{N}, i\neq  j \label{collision avoidance constraint}\\
&     &  & p_i(t, \theta_{i}(t)) \notin   \mathcal{O}_\mathrm{obs} \quad  \forall ~t,  \forall i \in \mathcal{N}          \label{collision avoidance obstacle}
\end{alignat}
\end{subequations}
       
with 
    \begin{equation}
p_i(t, \theta_{i}(t)) = p_i(t-1, \theta_{i}(t-1))+ \Delta p (\theta_i(t), v).
    \label{player position}
    \end{equation}
    
     The cost function $J(\boldsymbol{\theta_{i}})$ in \eqref{eq:optProb} is defined as
    \begin{equation}
        J(\boldsymbol{\theta_{i}}) = \Phi_\mathrm{goal}(\boldsymbol{\theta_{i}}) + \Phi_\mathrm{smooth}(\boldsymbol{\theta_{i}}) + \Phi_\mathrm{obs}(\boldsymbol{\theta_{i}}), 
        \label{overall_cost_function}
     \end{equation}
    where the three summands are defined as follows:

    \textit{(i)} The term $\Phi_\mathrm{goal}(\boldsymbol{\theta_{i}})$ tends to reduce the overall path length for each agent $i$ and, hence, models the goal-oriented attitude of the agent:
        \begin{equation}
        \begin{aligned}
            \Phi_\mathrm{goal}(\boldsymbol{\theta_{i}})=\sum_{t=1}^{T} \gamma(t) \| p_i(t, \theta_{i}(t))-p_i^{*} \|
        \end{aligned}
            \label{minimise_distance}
        \end{equation}
    with $\gamma(t)$ being a time-variant weight factor; $p_i(t, \theta_i(t))$ is the estimated position of agent \emph{i} at time $t$, considering a constant speed modulus $v$ and the heading control action $\theta_i(t)$ applied at time $t$, computed using
    the kinematic update Equation \eqref{player position}; and $p_i^{*}$ is the estimate of agent $i$'s goal in the time horizon $T$. In the absence of an explicit definition of a pedestrian's goal, we assume that, within the horizon $[t, t+T]$, the goal of agent $i$ lays on a straight line starting in $p_i(t)$ and oriented along the observed agent heading at time $t$.  Under these assumptions, the  practical meaning of the time horizon $T$ is the estimate of the time interval within which a pedestrian sets up and maintain their walking goal. 
    
    \textit{(ii)} The term $\Phi_\mathrm{smooth}(\boldsymbol{\theta_{i}})$ penalizes excessive rotations, thus promoting smooth trajectories. In fact, during navigation, humans tend to avoid too many changes of orientation to minimize their energy consumption \cite{mcneill2002energetics}: 
    \begin{equation}
        \begin{aligned}
         \Phi_\mathrm{smooth}(\boldsymbol{\theta_{i}})= \sum_{t=1}^{T} ( 1- \gamma(t) )  | \theta_i(t) - \theta_i(t-1) |
        \end{aligned}
            \label{smoothness}
     \end{equation}
     where $\theta_i(t)$, $\theta_i(t-1)$ are the orientation of the agent at time $t$ and $(t-1)$, respectively. We observe that the term $\Phi_\mathrm{smooth}(\boldsymbol{\theta_{i}})$ is weighted in a complementary fashion to $\Phi_\mathrm{goal}(\boldsymbol{\theta_{i}})$, to satisfy the assumption (further detailed in the Implementation section) of their  relative importance as long as the agent approaches its target.

    \textit{(iii)} The term $\Phi_\mathrm{obs}(\boldsymbol{\theta_{i}})$ tends to optimize the natural interaction with static objects. In fact, humans tend not to walk too close to static obstacles, unless it is necessary. For this reason, we model this behavior as a \emph{soft} constraint:
    
    \begin{equation}
        \begin{aligned}
         \Phi_\mathrm{obs}(\boldsymbol{\theta_{i}})= \sum_{t=1}^{T} \frac{\rho}{\| p_i(t, \theta_{i}(t))- p_{\text{obs}} \|}
        \end{aligned}
            \label{obst}
     \end{equation}
    where $\rho$ is a weighting factor and the denominator in \eqref{obst} is the distance between the agent position $p_i(t, \theta_{i}(t))$ and the closest \emph{static} obstacle $p_{\text{obs}}$ at time $t$. The exact procedure to compute $p_{\text{obs}}$ will be explained later. Practically, \eqref{obst} penalizes small distances between an agent and \emph{static} obstacles.

    The inequality in \eqref{collision avoidance constraint} is a hard constraint imposing to avoid other agents, assuming a circular region around agents as their vital space~\cite{hall1966hidden} to be avoided. 
    In this way, agent \emph{i} is required to maintain at least a minimum distance $\beta$ with other agents in the observer scenario. Constraint \eqref{collision avoidance obstacle} models the avoidance of static obstacles by imposing that the position $p_i(t,\theta_{i}(t))$ is outside the obstacle space $\mathcal{O}_\mathrm{obs}$, defined as a subset, possibly disconnected, of the 2D planar space, occupied by obstacles, where motion of agents is forbidden.

Equation~\eqref{player position} formalizes the kinematic update of the position of agent $i$ at time $t$, subject to a heading command $\theta_i(t)$, at a constant velocity $v$.

\begin{figure*}[h]

        \begin{subfigure}{0.4\textwidth}
         
          \includegraphics[width=\textwidth]{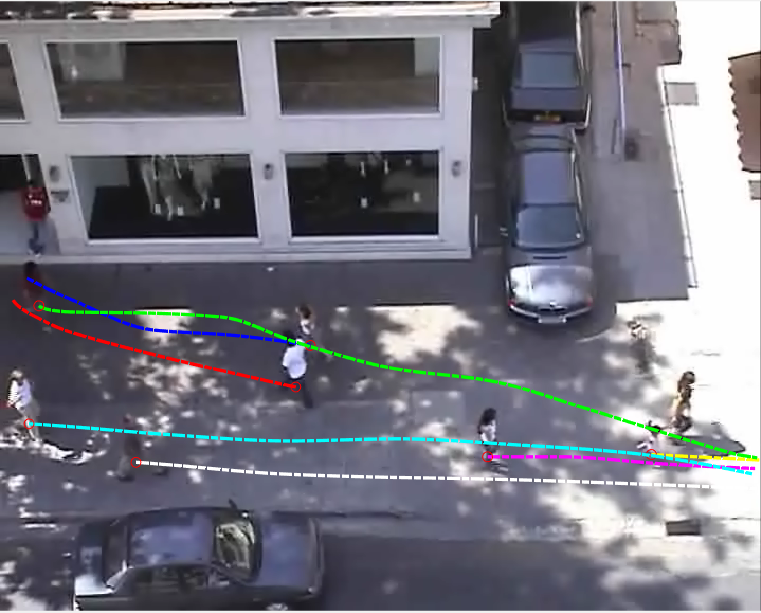} 
          \caption{}
          \label{scenario_1_real}
        \end{subfigure}
        \hfill
        \begin{subfigure}{0.4\textwidth}
          \includegraphics[width=\textwidth]{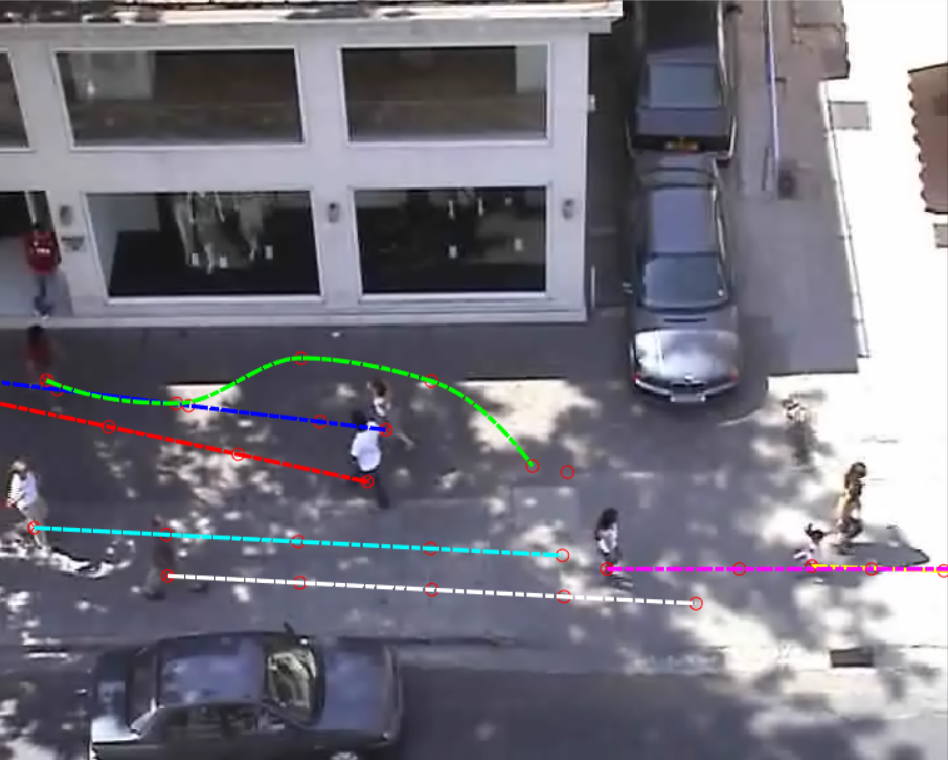} 
          \caption{}
          \label{scenario_1_model}
        \end{subfigure}
      
        \begin{subfigure}{0.4\textwidth}
    
          \includegraphics[width=\linewidth]{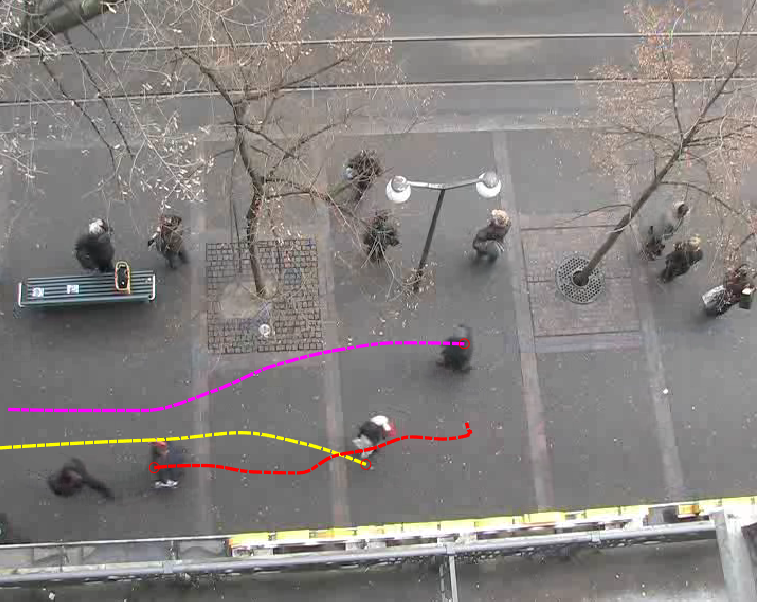}
          \caption{}
          \label{scenario_2_real}
        \end{subfigure}
        \hfill
        \begin{subfigure}{0.4\textwidth}
          \includegraphics[width=\linewidth]{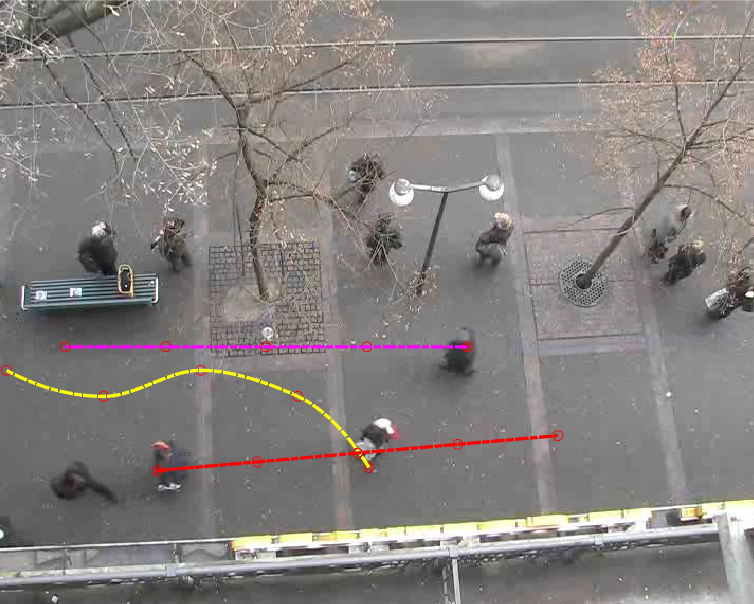}
          \caption{}
          \label{scenario_2_model}
        \end{subfigure}
        
        \caption{Validation of our human motion model based on game-theoretical approach with open-source surveillance videos~\cite{lerner2007crowds,pellegrini2009you}.\\
        \textbf{a)-c)} Real human trajectories; \textbf{b)-d)} Trajectories output of the game-theoretical model.}
        \label{Validation}

    \end{figure*}

\subparagraph{Validation}
The proposed game-theoretical human motion model is validated by conducting a qualitative comparison between generated trajectories and human ones, observed in open-source surveillance videos~\cite{lerner2007crowds,pellegrini2009you}. 
These surveillance videos, used to validate the proposed model, show a typical urban scenario in which multiple agents walk interacting with each other and avoiding static obstacles. Figure~\ref{Validation} illustrates the frames, randomly selected, of the surveillance videos of two different scenarios. Specifically, Figure~\ref{Validation} compares real trajectories executed by humans (Figs.~\ref{scenario_1_real} and~\ref{scenario_2_real}) with the estimated trajectories generated  for all agents by the proposed model solving our game-theoretical problem (Figs.~\ref{scenario_1_model} and~\ref{scenario_2_model}).

We observe that, in both the illustrated scenarios, our game-theoretical approach generates collision-free trajectories (Figs.~\ref{scenario_1_model} and~\ref{scenario_2_model}) that are smooth and resemble those executed by their human counterparts. However, we note that the trajectories generated by our algorithm exhibit a sharper reaction than humans in the vicinity of  surrounding agents. This is evident while comparing  Figure~\ref{scenario_1_real} and Figure~\ref{scenario_1_model}, focusing on the interaction between the green trajectory and the blue one. A comparable circumstance can be observed in Figure~\ref{scenario_2_real} and Figure~\ref{scenario_2_model}, with reference to the yellow trajectory.
This phenomenon is most likely caused by the discrete action set associated with each agent. Notably, in our implementation an agent can choose one out of seven  possible headings inside their own visibility zone, resulting in a resolution of $ \pm \sfrac{\pi}{6}~\mathrm{rad}$, in the attempt of  minimizing the corresponding cost function. On the other hand, human subjects can select their heading over an infinite set. 

A further cause of discrepancy between human and game-theoretical trajectories resides in the kinematic update of the agent position in Equation~\eqref{player position} --a linear update with constant heading and velocity over the whole sampling step-- and the estimation of the human target, assumed to be constant over an interval of duration $T$ --actually an unknown, subject to the very stochastic nature of human behavior.

  \subparagraph*{Algorithm}
 The game-theoretical model of pedestrian motion described above is used to inform a robotic trajectory planner for autonomous robots moving in populated environments.

\begin{algorithm}[H]
\textbf{Initialization}:\\
$p_{\mathrm{robot}} \leftarrow \mathrm{InitializeRobotPosition}$\\

\Repeat{$p_{\mathrm{robot}} = p_{\mathrm{goal}}$}{

 $\mathrm{GroupRecognition}$\\
 $ \boldsymbol{\theta} \leftarrow \mathrm{FirstEstimation}$ \\
 
 \ForEach{agent $i$ (robot included)}{
 $C_\mathrm{obs} \leftarrow false$ [flag defining the collision with obstacles for agent $i$] \\
$C_\mathrm{agents} \leftarrow false$ [flag defining the collision between agents for agent $i$]\\
 $ C_\mathrm{obs}, C_\mathrm{agents}  \leftarrow \mathrm{CheckCollision}(i, \boldsymbol{\theta})$ \\
 
  $ \boldsymbol{\theta_i} \leftarrow \mathrm{ComputeSolution}(i, \boldsymbol{\theta}, C_\mathrm{obs}, C_\mathrm{agents})$\\
   }

   $ p_{\mathrm{robot}} \leftarrow \mathrm{UpdateRobotPosition}$ 
   
   }

 \caption{Main algorithm}
\label{Algorithm_main}
\end{algorithm}

\vspace{12pt}

\begin{algorithm}
	\SetKwFunction{main}{$\mathrm{ComputeSolution}$}
	\SetKwProg{Fn}{}{}{}
	\Fn{\main{$\boldsymbol{i, \theta}, C_\mathrm{obs}, C_\mathrm{agents}$}}
	{
		\eIf{$C_\mathrm{agents}$}{
		
            $\boldsymbol{\theta_i}^\mathrm{gt} \leftarrow \mathrm{GTPlanner}(\boldsymbol{\theta}) $ [Solution to Algorithm \ref{Algorithm_2}]\\
            $\boldsymbol{\theta_i}^\mathrm{dec} \leftarrow \mathrm{Decelerate} $\\
            \eIf{$\mathrm{Cost}(\boldsymbol{\theta_i}^\mathrm{gt}) \leq \mathrm{Cost}(\boldsymbol{\theta_i}^\mathrm{dec})$}
            {
                $\boldsymbol{\theta_i} \leftarrow \boldsymbol{\theta_i}^\mathrm{gt}$
            }
            {
                $\boldsymbol{\theta_i} \leftarrow \boldsymbol{\theta_i}^\mathrm{dec}$
            }
            
        }
        {\If{$C_\mathrm{obs}$}{
            $\boldsymbol{\theta_i} \leftarrow \mathrm{IndividualOptimization}$
        }}
    \Return $\boldsymbol{\theta_i}$
	}
	\caption{The $\mathrm{ComputeSolution}$ function}
	\label{Model_solution_algorithm}
\end{algorithm}

The main steps executed by the proposed trajectory planner are described in Algorithm \ref{Algorithm_main}.
First, the robot position ($p_{\mathrm{robot}}$) is initialized using the function $\mathrm{InitializeRobotPosition}$.
Then, the algorithm executes an  iterative procedure that stops when the robot reaches its target position ($p_{\mathrm{goal}}$). Here, we will refer to both humans and the robot with the term ``agent''. Each iteration performs five main steps:
recognition of groups of humans  ($\mathrm{GroupRecognition}$), first estimation of trajectories for all agents ($\mathrm{FirstEstimation}$), collision checking between agents and with obstacles ($\mathrm{CheckCollision}$), computation of the agent  trajectory ($\mathrm{ComputeSolution}$), and update of the robot position using the computed trajectory ($\mathrm{UpdateRobotPosition}$).
This iterative procedure predicts the agents' motion and generates the robot optimal trajectory over the fixed time horizon $T$, by applying the strategy detailed below. After such an optimal trajectory for the robot is computed, only the action corresponding to the first time step is actually applied to the robot and the process is repeated until the robot reaches its goal.

In the following, each step of the Algorithm \ref{Algorithm_main} is detailed:

\begin{itemize}

    \item $\mathrm{GroupRecognition}$. The algorithm performs the \emph{group recognition} of agents considering the observed orientation of each agent, and the distances between them. In fact, a group is typically moving maintaining a common orientation and keeping a  distance between agents shorter than the vital space typical of the single agent.
    Upon recognition, groups are considered as \emph{unique entities} and treated as single agents in the subsequent phases. 
    
    \item $\mathrm{FirstEstimation}$. A preliminary estimation of all agents' trajectories (i.e., $\theta$) is performed, projecting hypothetical rectilinear trajectories over the interval $T$.

    \item $\mathrm{CheckCollision}$. Given the trajectories of all agents ($\theta$), previously estimated by the $\mathrm{FirstEstimation}$, the $\mathrm{CheckCollision}$ function detects the possible occurrence of collisions between an agent $i$ with obstacles and other agents, activating the flag variables $C_\mathrm{obs}$ and $C_\mathrm{agents}$, respectively.
    In particular, we refer to the occurrence of a collision each time the individual vital space of an agent is violated.

    \item $\mathrm{ComputeSolution}$. Considering the estimated trajectories ($\theta$), and the flags $C_\mathrm{obs}$ and $C_\mathrm{obs}$, Algorithm \ref{Model_solution_algorithm} computes a solution of the motion planning problem for an agent $i$ selecting one of the possible cases:
    \begin{enumerate}[(i).]
        \item if a collision with other agents is envisaged, two alternative solutions are evaluated. Hence, the solution that involves the lowest cost of Equation \eqref{overall_cost_function} will be selected. 
            
        The first solution ($\boldsymbol{\theta_i}^\mathrm{gt}$) is computed using the strategy defined in Algorithm \ref{Algorithm_2}, where trajectories are generated seeking for a Nash equilibrium solution of the game presented in the \emph{Game description} section. 
            
        The second solution is computed through the $\mathrm{Decelerate}$ function, which evaluates the opportunity  to \emph{decelerate} --a typical human behavioral trait in navigation-- to avoid the collision with other agents.
        In particular, after identifying the discrete time step $t$ at which a collision  between  agent $i$ and other agents is envisaged to occur, the cost associated with sixteen different deceleration patterns is evaluated using the cost function  \eqref{overall_cost_function}, provided that  constraints in Equations \eqref{collision avoidance constraint} and \eqref{collision avoidance obstacle} are satisfied; 
        
        \item if an agent is envisaged to collide  with a \emph{static} obstacle ($C_\mathrm{obs}$), the agent   solves its individual  optimization problem described above (without playing the game and, hence, not seeking for the Nash Equilibrium);  
        
        \item if no collision between agents or static obstacles is envisaged, trajectories are kept linear, maintaining the current heading and constant velocity, practically implementing what was already computed in the $\mathrm{FirstEstimation}$ procedure.

        \end{enumerate}
        
      \item $\mathrm{UpdateRobotPosition}$. Considering the computed trajectory of the robot, the action corresponding to the first time step is executed and the  robot position is updated using Equation~\eqref{player position}.
      
\end{itemize}

\begin{algorithm}[H]
\SetAlgoLined
\textbf{Initialization}:\\
     \qquad $k \leftarrow 1$ [iteration index] \\
     \qquad $ \boldsymbol{\theta}^k  \leftarrow \boldsymbol{0} $ [straight paths for all agents as $\mathrm{FirstEstimation}$] \\
     \qquad $ i \leftarrow 1 $ [agent index] \\

\textbf{Iterate until convergence}:

 \qquad $ \bar{p}_j^k \leftarrow $ \eqref{player position}, given $\boldsymbol{\theta}^k$, for all $j$ 
 [present and future predicted positions of all agents]\\

 \qquad $\boldsymbol{\theta_i^{k+1}} \leftarrow  $ solution to Eq. (\ref{eq:optProb}--\ref{collision avoidance obstacle}), given $ \left( \bar{p}_j^k \right)_{j \neq i}$ 
 [best response to all other agents] \\

\eIf{$i < \mathcal{N}$}{
   $i \leftarrow i + 1 $ [move on to next agent]
   }{
   $i \leftarrow 1 $, $ k \leftarrow k + 1 $ [move on to next iteration]
  }

 \caption{Nash trajectory computation}
\label{Algorithm_2}
\end{algorithm}

\subparagraph*{Implementation} 

The algorithm presented above has been implemented in Matlab and the main implementation choices are discussed in what follows.

The discrete time step has been set to $\Delta t = 1.2 \mbox{s}$. The time horizon for optimization has been set to $T = 4$, that is, 4.8 seconds. In the main paper, we opted to keep a unitary discrete time step, to enhance readability.

As previously stated, each agent can execute actions taken from an action set $\Theta$ of finite size. Specifically, in our implementation, each agent has seven possible actions for $\theta_i(t)$, which represents the heading within the agent \emph{visibility} zone. Namely, $\theta_i(t)$ is updated as   $\theta_i(t) = \theta_i(t-1) + u(t-1)$, where $u(t-1)$ takes values in the finite set $\Theta=\{ -\sfrac{\pi}{2}, -\sfrac{\pi}{3}, -\sfrac{\pi}{6}, 0, \sfrac{\pi}{6}, \sfrac{\pi}{3}, \sfrac{\pi}{2} \}~\mathrm{rad} $. We remark that we limited the cardinality of $\Theta$ to seven, pursuing a trade-off between satisfactory performance and reasonable computational complexity of the algorithm.

In Equation \eqref{collision avoidance constraint}, the $\beta$ parameter is set considering the Hall convention \cite{hall1966hidden} that posits the existence of a vital space of circular shape that ensures comfort conditions for human navigation. The value of $\beta$ has been estimated through the analysis of the open-source surveillance videos~\cite{lerner2007crowds,pellegrini2009you}.

In Equation \eqref{overall_cost_function}, the term ($\Phi_\mathrm{obs}(\boldsymbol{\theta_{i}})$) can be neglected if the first estimation of the agent trajectory does not intersect any static obstacle. Otherwise,  $\Phi_\mathrm{obs}(\boldsymbol{\theta_{i}})$ in Equation \eqref{obst} is computed referring to the closest obstacle, toward which the agent is projected to collide. 
Then, the closest point of such  obstacle to the agent position is computed ($p_{\text{obs}}$). To reduce the computational load, obstacles are mapped into a discrete spatial map overlapping with the 2D environment. The map consists of a rectangular matrix of 576x720 cells, which are marked as being occupied by an obstacle or free from them. Each cell covers approximately a square of 1.8x1.8 cm.  

The weight $\gamma(t)$
in Equations \eqref{minimise_distance} and \eqref{smoothness} is selected as a time-varying term that is used to balance the relative importance of terms $\Phi_{\mathrm{goal}}(\boldsymbol{\theta}_i)$ and $\Phi_{\mathrm{smooth}}(\boldsymbol{\theta}_i)$ over the optimization horizon $T$. This choice emerges from the analysis of the available surveillance videos, where we observed that the minimization of the distance to goal typically prevails on the smoothness requirement as long as the agent gets closer to their goal, and vice versa. Considering $T=4$ time steps, we chose the following sequence for $\gamma(t)$, starting from a generic time instant $t^*$: $\gamma(t^*)=0.6$, $\gamma(t^*+1)=0.7$, $\gamma(t^*+2)=0.8$, $\gamma(t^*+3)=1.0$.

        \subsection*{Questionnaire and a-priori power analysis}
    \subparagraph*{Questionnaire}
   \label{Experimental setup}
    The proposed methodology is validated using a variation of the Turing test \cite{saygin2000turing}, which evaluates whether  the robot behavior, controlled by the game-theoretical method, is comparable to or indistinguishable from  human navigation patterns.
    
\begin{figure*}[h]
        \centering
        \begin{subfigure}{0.4\textwidth}
         
          \includegraphics[width=\textwidth]{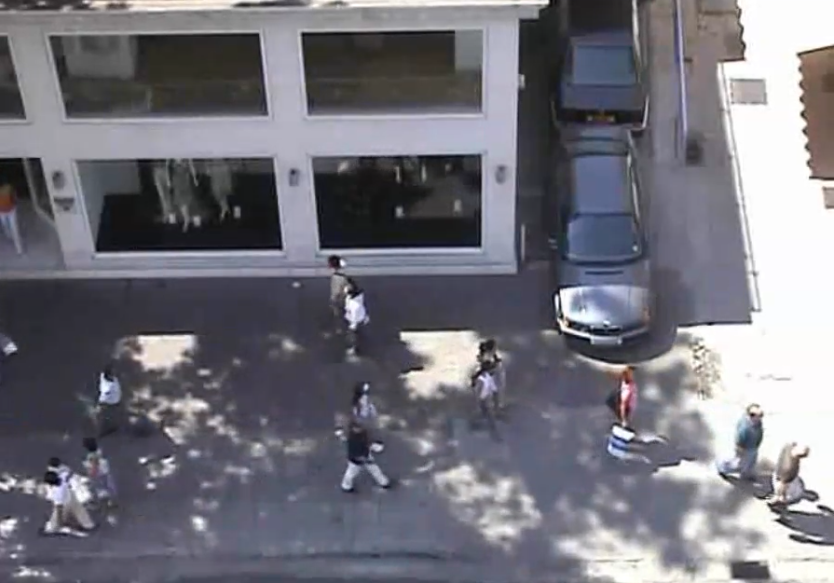}
          \caption{}
          \label{training1}
        \end{subfigure}
        \hfill
        \begin{subfigure}{0.4\textwidth}
        
          \includegraphics[width=\textwidth]{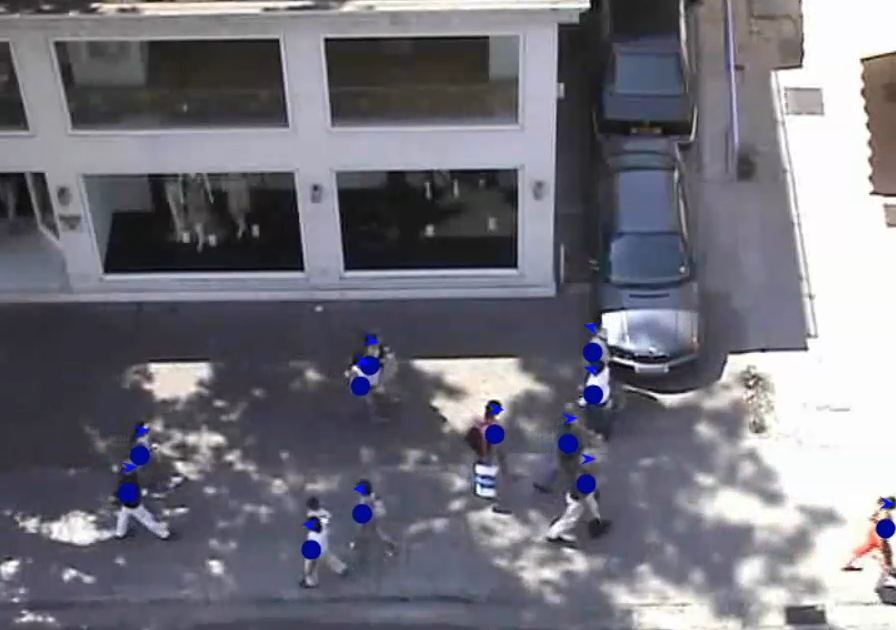}  
          \caption{}
          \label{training2}
        \end{subfigure}
      
        \begin{subfigure}{0.4\textwidth}
    
          \includegraphics[width=1.1\textwidth]{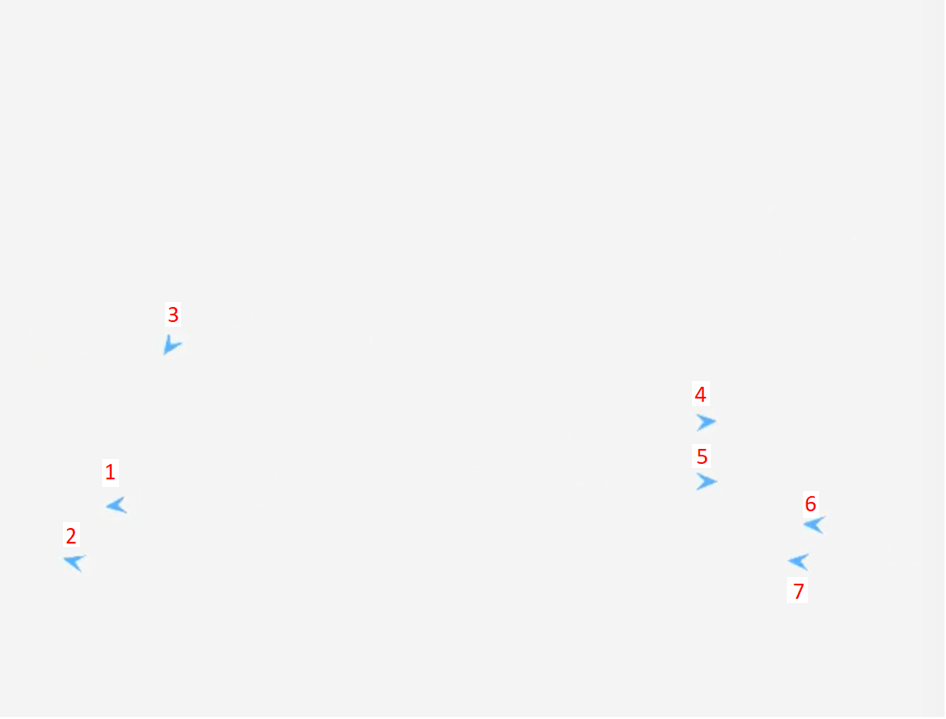} 
          \caption{}
          \label{first section}
        \end{subfigure}
        \hfill
        \begin{subfigure}{0.4\textwidth}
          \includegraphics[width=1.1\textwidth]{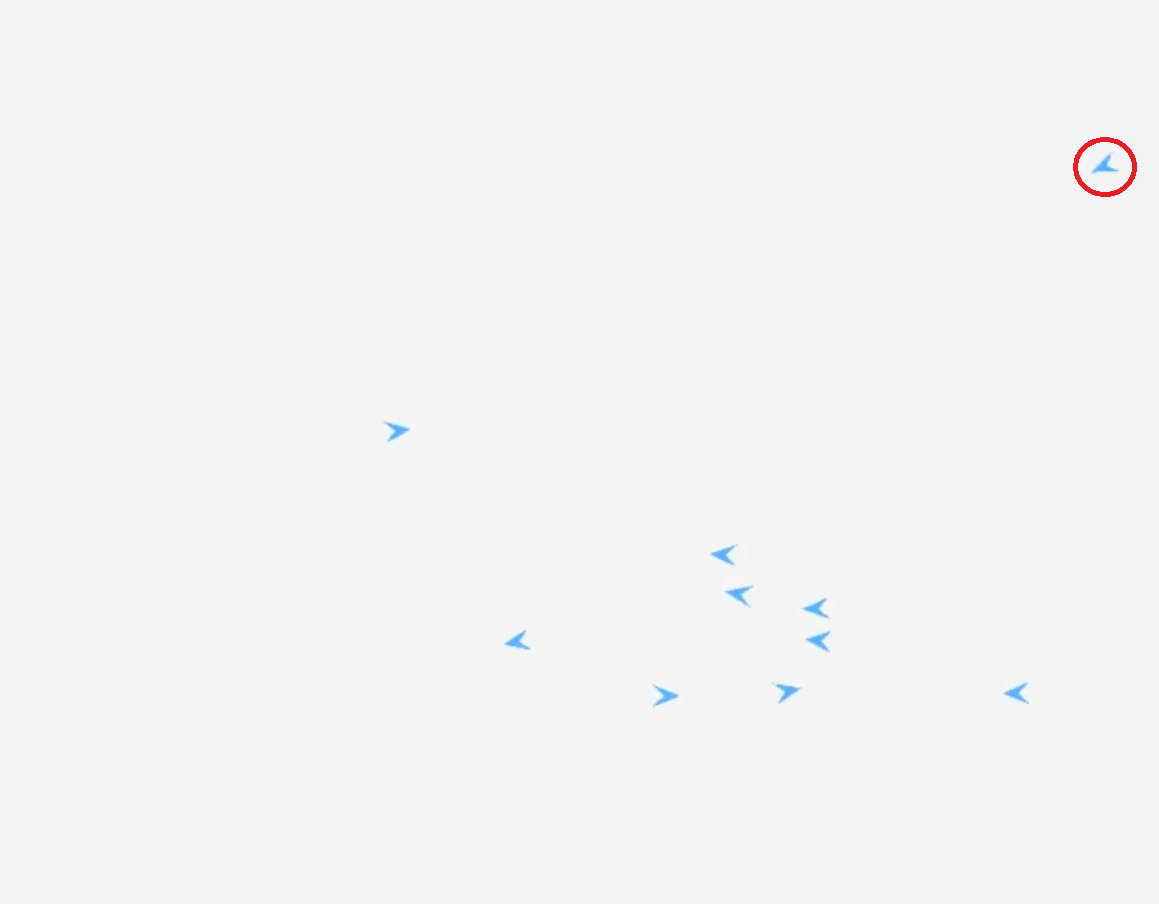}
          \caption{}
          \label{second section}
        \end{subfigure}
        
        \caption{Overview of the survey questionnaire.\\
        \textbf{a)} Training part with open-source surveillance video \cite{lerner2007crowds}; \textbf{b)} Training part, intermediate scenario; \textbf{c)} Second part of the test, i.e. recognizing the motion of the "weird" arrow in the videos, if any; \textbf{d)} Third part of the test, i.e. follow the circled arrow.}
        \label{online test}

    \end{figure*}

    The variation of the Turing test consists of an online questionnaire composed by three main parts: a preliminary training part and two different parts to collect data. A screen for each part of the survey questionnaire is illustrated in Figure~\ref{online test}.
    The survey questionnaire starts with general questions about the participant, such as gender, age, and the level of professional experience in robotics field on a Likert scale~\cite{likert1932technique} from 1 (no experience)  to 5 (expert). Then, a preliminary training part allows the participant to become familiar with the working environment (Figures~\ref{training1} and~\ref{training2}).
    Training videos show pedestrians moving in an urban environment, as shown in Figure~\ref{training1}. In particular, the training part guides the participant from a typical urban scenario of Figure~\ref{training1} to the particular scenario used in the other parts of the test illustrated in Figure~\ref{first section}. The intermediate scenario of Figure~\ref{training2} is designed to gradually guide the participant to the final set-up. 
    
    In the testing scenario of Figure~\ref{first section}, agents (pedestrians and robot) have been replaced with arrows and the urban environment has been removed to prevent the participant from focusing on the scenario, rather than on the movement of agents.
    
    In the second part, the participant watches 21 videos randomly (about 15 seconds each) consisting in three different experimental conditions: 7 videos show an environment with only pedestrians; other 7 videos a scenario with pedestrians and a robot controlled with an algorithm at the state-of-the-art (the Enhanced Vector Field Histogram~\cite{ulrich1998vfh+}); and the remaining 7 videos show a scenario with pedestrians and a robot controlled with the proposed algorithm. In all experimental conditions, robot trajectories are re-planned with a frequency of 2 Hz.
    
    At the end of each video, the participant affirms if they have noticed an arrow that moved with a \emph{weird} motion. In the affirmative case, they point that arrow, as shown in Figure~\ref{first section}.
    
    In the last part of the survey questionnaire, the participant watches the same 21 videos of the previous part but in a different random order.

   Unlike the previous part, an arrow is circled in red, as shown in Figure~\ref{second section}. Then, the participant have to determine if the selected arrow is a real pedestrian and, then, they evaluate the \textit{naturalness} of the arrow motion on a Likert scale~\cite{likert1932technique} defined in a range from 1 (completely unnatural) to 5 (completely natural).
    
    All videos used in the survey questionnaire are generated from an open-access dataset~\cite{lerner2007crowds}.
    
    The test takes about 20 minutes to be completed properly.
    The test has three \emph{rules}: \textit{(i)} the participant cannot pause the video; \textit{(ii)} the participant can watch videos only once; \textit{(iii)} the participant should complete the test without interruptions or distractions.

\subparagraph*{A-priori power analysis}
 Preliminary, we conducted an a-priori power analysis to estimate the number of participants required to provide acceptable and significant statistical results \cite{prajapati2010sample}. To this aim, we used the free software G*Power \cite{erdfelder1996gpower}.
First, we identified our case analysis as a non-parametric study, since non-parametric statistical tests make no constraints and prerequisites on the data distributions \cite{corder2014nonparametric}. Then, we assumed that the data collected after the a-priori study would be analyzed via the non-parametric Kruskal-Wallis test because our \emph{independent} variables have more than two independent groups (HO, GT, and VFH) and our \emph{dependent} variables (the rating of the weirdness motion, human-likeness, and naturalness of movement) are ordinal.

Based on \cite{prajapati2010sample}, we computed the total sample size considering the ANOVA test \cite{roberts2014student}, i.e., the parametric-equivalent test of the Kruskal--Wallis one and then multiplied the result by the corrective factor ARE, obtaining the equivalent sample size of the non-parametric Kruskal--Wallis test.
The result of the a-priori analysis for our non-parametric test is about $152$ volunteers, considering an alpha level equal to $5\%$, power of the study $80\%$ and the three number of groups, corresponding to the three different experimental conditions.
We recruited the participants using the Institutional mail of Politecnico di Torino and then we distributed an online questionnaire to students and university staff. Ultimately, we collected $691$ responses, exceeding the sample size of 152.

\section*{Results}

\subsection*{A game-theoretical framework for social acceptability of robotic trajectories}

Figure~\ref{Fig:Intro_Image} schematizes the proposed procedure for the realization and validation of our game-theoretical framework for the social acceptability of robotic trajectories. The methodology can be subdivided in four main logical phases, corresponding to panels in the figure. First, a game-theoretical model of pedestrian motion is devised an its parameters are tuned on the basis of the analysis of human motion videos (panel (a)). Second, a robotic trajectory planner informed by the game-theoretical pedestrian model is realized. The robot is deployed and operated in a virtual humanly populated environment, where humans execute real trajectories extracted from videos. In this phase, three important performance metrics in robotic trajectory planning are evaluated and compared with the state-of-the-art VFH algorithm (panel (b)). Third, the virtual environments containing humans and the robot are processed and prepared to be administered for the validation questionnaire (panel (c)). Finally, the questionnaire is administered and the results are collected and analyzed (panel (d)). 

\subsection*{Analysis of performance metrics}

We performed a preliminary assessment of the trajectories generated in three experimental conditions, which differ for the algorithm governing the motion of a selected agent (i.e., either a robot or a human being): in the condition \emph{humans only} (HO), all the agents were human beings moving in a real environment; in the condition \emph{humans and GT} (GT), one of the agents was controlled by our game-theoretical  algorithm, while the other agents were human beings; and in the condition \emph{humans and VFH} (VFH), one of the agents was controlled by the VFH algorithm \cite{ulrich1998vfh+}, and the other were human beings. Each experimental condition comprises seven different experiments, differing for the quantity of human subjects and their motion patterns. 

The virtualized environment is constructed by processing movies collected from surveillance cameras of populated environments ~\cite{lerner2007crowds}, obtaining a 2D arena where virtual agents reproduce the human motion captured in the video. In the HO condition, the performance metrics are evaluated in the original arena, with reference to a randomly selected human being. In the GT and VFH conditions, a virtual agent is introduced in the arena and commanded to navigate through the existing virtual agents (corresponding to human beings) using the given trajectory planner. 

Three widely adopted metrics, deemed as important for socially navigating robots, were evaluated across the three experimental conditions: the Path Length Ratio (PLR), the Path Regularity (PR), and the Closest Pedestrian Distance (CPD)~\cite{biswas2021socnavbench}.

\begin{figure}[h]
         \centering
        \includegraphics[width=0.8 \textwidth] 
         {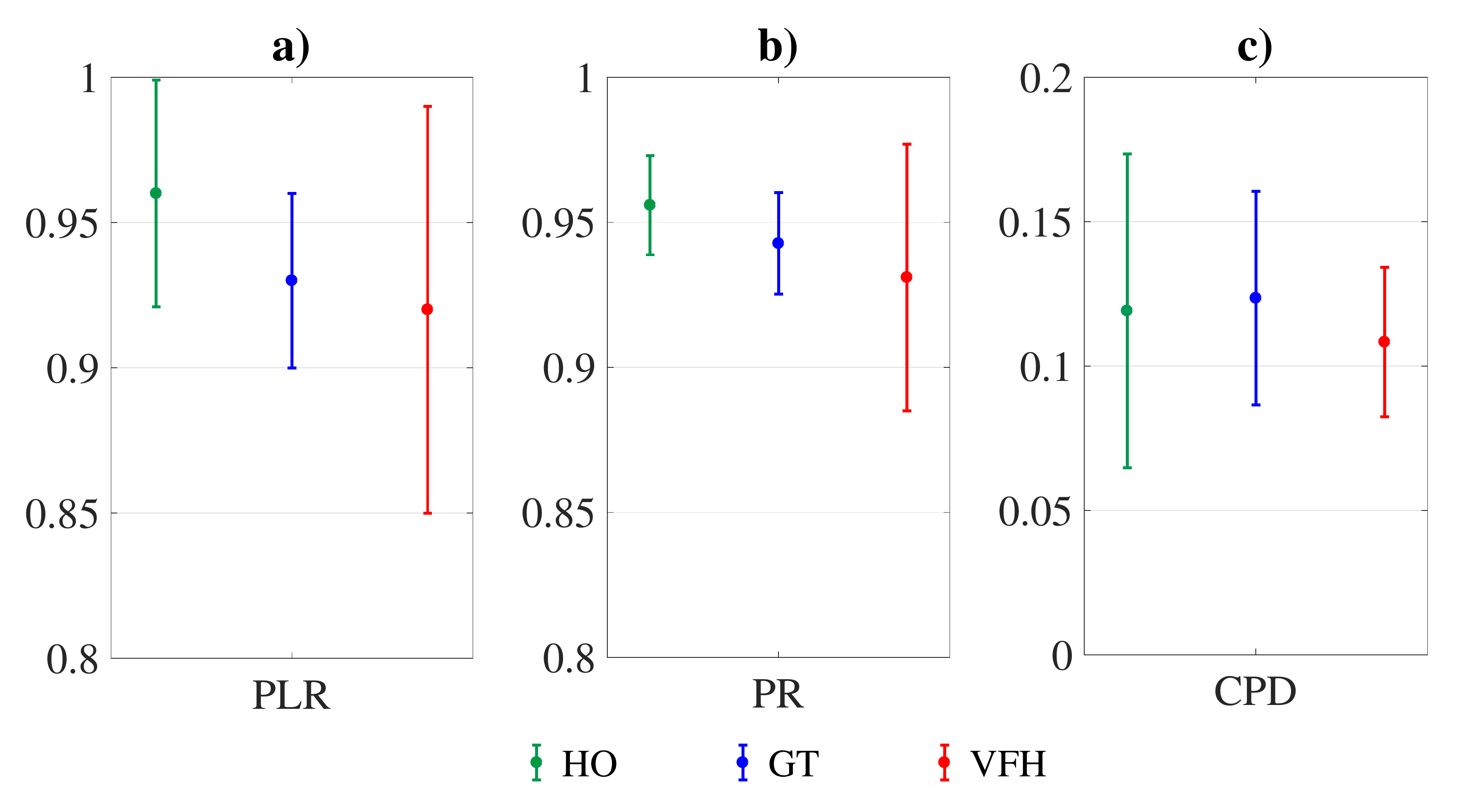}
        \caption[width=1\textwidth]{Metrics parameters used to evaluate the social trajectories in the three groups of the experimental conditions. HO: video with human only; GT: video with humans and a robot driven by a game-theoretical trajectory planner; VFH: video with humans and a robot driven by a vector field histogram algorithm. \textbf{a)} PLR (Path Length Ratio) is the average of the ratio of the line of sight path distance between the start and goal and the real agent's trajectory; \textbf{b)} PR (Path Regularity) is the average regularity of the agent's trajectory
        ; \textbf{c)} CPD (Closest Pedestrian Distance) is the average distance between the agent and the closest pedestrian.  
        } 
         \label{Grafico_parameters}
       
    \end{figure}

The PLR is defined as the ratio between the length of the line-of-sight path between the initial and final point of a path  and the actual path length between the same two points~\cite{biswas2021socnavbench}. A higher path length ratio is usually preferred, since it indicates that an agent minimizes the length of the path to reach its goal. We computed the PLR for each experiment and we illustrate its average values across the three experimental conditions in Figure~\ref{Grafico_parameters}a.  While the results in Figure~\ref{Grafico_parameters}a suggest that the HO scenario was characterized by the highest average PLR, followed by GT and VFH.
     
The PR quantifies to what extent a path is similar to a straight line~\cite{biswas2021socnavbench}. Following normalization, $PR=1$ corresponds to a straight path from start to goal. Values of PR closer to one are preferable, since they are indicative of a smoother motion, without excessive changes of direction. In Figure~\ref{Grafico_parameters}b, the average PR for each experimental condition is illustrated, where the highest average value pertains to HO, followed by GT and VFH. These results appear in line with the tenet that humans tend to minimize their energy, thus avoiding sudden changes of orientation, and with the design principle of the VFH algorithm, which avoids obstacles only when the agent is close to them~\cite{ulrich1998vfh+}, entailing swift changes of orientation to get away from them. 
     
The CPD is defined as the distance from the closest pedestrian, normalized with respect to the maximum length measurable during experiments, which is the diagonal of the experimental arena. Also for this parameter, the attainment of values closer to one is desirable, as this implies a good tendency in staying clear from humans when following planned trajectories. Average values of CPD in the three experimental conditions are illustrated in Figure~\ref{Grafico_parameters}c, where the highest average value is related to  GT, followed by HO and VFH. The reason for the latter is presumably due to the purely reactive design of the VFH algorithm.  We posit that the intermediate ranking of HO with respect to CPD is due to  the ability of humans to evaluate situations on a case-by-case basis. 
     
While the rankings described above are suggestive of superior performance metrics attained by GT over VFH, the verification of the statistical significance of these comparisons is in order. To this aim, Kruskal-Wallis analysis~\cite{ostertagova2014methodology} was executed across the three metrics, revealing the non-achievement of significant statistical distinguishability ($p=0.286$ for PLR, $p=0.400$ for PR, $p=0.834$ for CPD). The reason behind such observations is strictly related to the consideration of only seven experiments for each experimental condition, with differential degree of variability, and thus characterized by a limited statistical power. In line with these considerations, we conducted a systematic analysis of the inter-experiment variability within each experimental condition (Levene’s test ~\cite{gastwirth2009impact}). Essentially, we evaluated the extent to which each algorithm generated experiments that were similar to one another. Such an analysis revealed that there exists a significant differential variability with respect to PR ($F_{2,18}=3.75$, $p = 0.043$). Thus, we performed a post-hoc analysis that revealed much more variability in the VFH videos compared to HO and GT (VFH vs. HO: $p = 0.038$; VFH vs. GT: $p = 0.040$; HO vs. GT: $p = 0.97$); similarly, albeit not statistically significant, we observed a trend toward increased variability in VFH experiments with respect to PLR ($F_{2,18}=3.22$, $p = 0.064$). Finally, the inter-experiment variability within each experimental condition was indistinguishable concerning CPD ($F_{2,18}=2.31$, $p = 0.130$). These results indicate that, albeit indistinguishable in absolute values, the reproducibility and predictability of each experimental condition in terms of PR and PLR was much higher in HO and GT than in VFH scenario.

\subsection*{Survey questionnaire}
We collected 691 responses to the survey questionnaire, where participants were in majority men in their thirties with very little experience on robotics (Table~\ref{First part data}). 

\captionsetup[table]{labelfont={bf}}
 \begin{table}
     
    \begin{center}
\begin{tabular}{ |cc| } 
 \hline
 Number of participants  &  691 \\ 
 \hline
 Gender & 58\% male and 42\% female \\
 \hline
 Age & 29.44$\pm$11.30 \\ 
 \hline
 Experience with robotics & 1.5$\pm$0.86\\
 \hline

\end{tabular}
\end{center}
     \caption{Demographic characteristics and experience with robotics on a scale from 1 (minimum experience) to 5 (maximum experience) collected during the first part of the test.}
     \label{First part data}
 \end{table}

A power analysis~\cite{prajapati2010sample} indicated that the adequate statistical power was guaranteed with 152 participants. Since the number of participants largely exceeded the required sample size, we opted for a bootstrapping approach~\cite{efron1994introduction},  in which we randomly sampled 152 observations from the complete pool of responses and iterated this process 100 times. Adopting this procedure, we kept the sample size to the appropriate number (thus reducing the odds to obtain biologically irrelevant findings~\cite{johnson1999insignificance}) and increased the generalizability of our findings by testing their robustness against repeated observations. 

  The test was composed by three parts: (i) in the first part, the participant underwent a training phase to become familiar with the working environment  (see Figure~\ref{online test}a-b); (ii) in the second part, the participant watched $21$ videos reproducing the experiments in the three experimental conditions, where both the background and the agents are concealed --blue arrows over a gray background-- (Figure~\ref{online test}c illustrates a single experiment); (iii) in the third and final part, the participant watched the same $21$ videos, where they were asked in addition to focus on a circled arrow (Figure~\ref{online test}d illustrates a single experiment). The participant is unaware that the circled arrow targeted a random human agent in the HO condition and the robotic agent in GT and VFH conditions.
  
The execution of each part entails answering specific questions. In the first part of the survey questionnaire, the participants were required to provide their gender, age, and level of experience in robotics. To assess the level of social acceptance of our game-theoretical trajectories, in the second part  (following habituation), we asked the participants to say if they perceived  “weirdness” in the motion observed in the videos, and then to indicate which is the perceived “weird” arrow, if any. In the third part, participants were requested to determine whether the circled arrow is a human or not. Then, participants were asked to rate the naturalness of the motion of the circled arrow on a Likert scale.

Experimental outcomes were analyzed with the Kruskal-Wallis test, as the independent variables belong to more than two independent groups (HO, GT, and VFH) and the dependent variables (the rating of the weirdness of motion, the classification as a human or not, and the naturalness of movement) are ordinal~\cite{corder2014nonparametric}. Our null hypothesis, $H_{0}$, posits that all experimental conditions (HO, VFH, GT) are perceived as indistinguishable. To statistically reject the $H_{0}$ hypothesis and understand if there exist differences among  experimental conditions, we computed the \emph{p} value considering $5$\% of significance level~\cite{kraska2013nonparametric}. Then, we performed a Bonferroni post-hoc analysis~\cite{foster2018introduction} to determine which groups are classified as significantly different from each other. This procedure was adopted for all the questions in the second and third part of the survey questionnaire, except for the second question of the second part, where participants were asked to indicate the perceived ``weird'' arrow, if any. In this case, the answers expressed relative the HO scenario were discarded, since all arrows corresponded to human beings and an indication of weirdness would not make sense to our research question. As a consequence, only to experimental conditions had to be compared (GT and VFH) and, to this aim, we used the Mann--Whitney test~\cite{mann1947test,fay2010wilcoxon}. 

\begin{figure}[h]
    
        \centering
      \includegraphics[width=0.8\textwidth] 
       {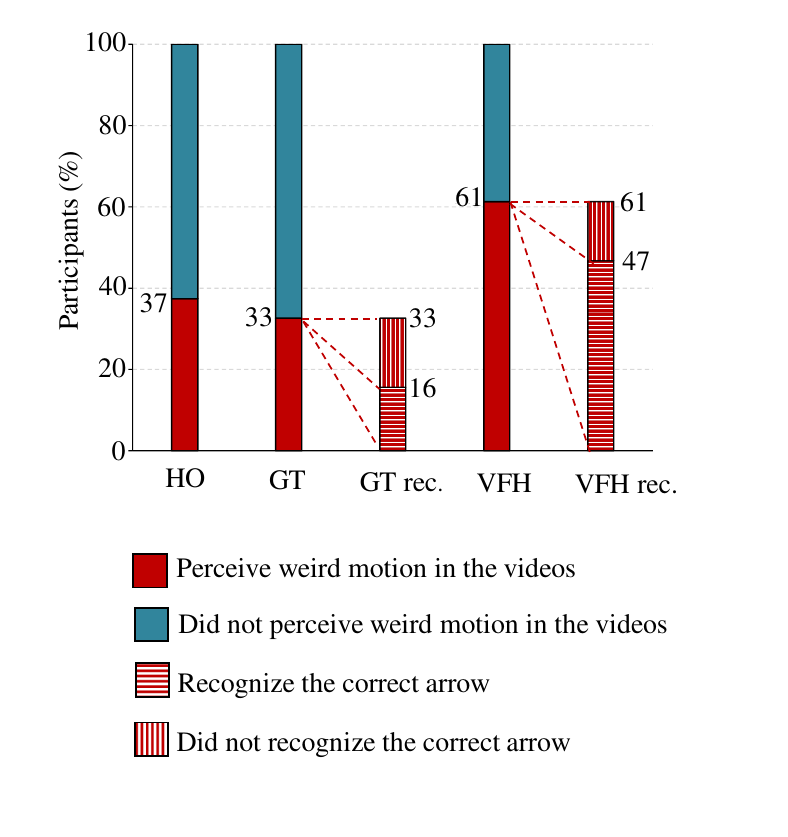} 
        \caption{Result of the second part of the survey questionnaire considering the participants that perceive a ``weird'' motion in the 3 groups of experimental conditions and recognize the correct arrow (the robot) in the populated environment (GT rec. , VFH rec.). HO: video with humans only; GT: video with humans and a robot driven by a game-theoretical trajectory planner; VFH: video with humans and a robot driven by a vector field histogram algorithm.}
    \label{part_2_analysis}
       
    \end{figure}

\begin{figure}[h]
         \centering
        \includegraphics[width=0.8 \textwidth] 
       {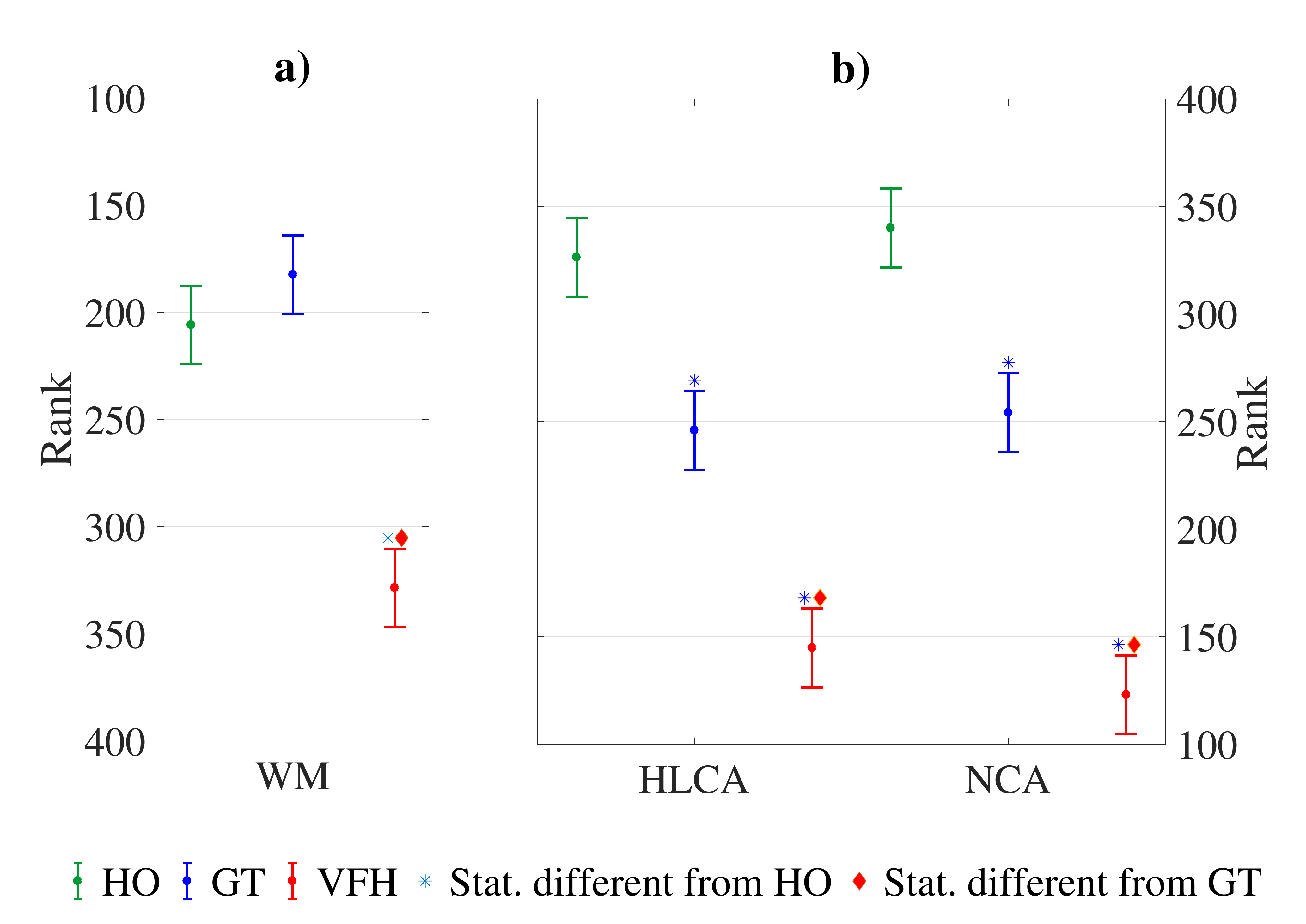}
        \caption[width=1\textwidth]{Summary of the post-hoc Kruskal-Wallis test of the survey questionnaire. The mean rank of each group is plotted for each part of the test and the comparison interval is highlighted in order to easily get the significant difference between the three experimental conditions. \textbf{a)} Second part of the test in which the attention of the participant is not focused on one arrow in particular. WM: weirdness motion; \textbf{b)} Third part of the test in which the participant is focused on the circled arrow.  HLCA: human-likeness of the circled arrow; NCA: naturalness of the circled arrow. 
       
        HO: video with humans only; GT: video with humans and a robot driven by a game-theoretical trajectory planner; VFH: video with humans and a robot driven by a  Vector Field Histogram algorithm.
        The blue asterisk highlights the statistical difference from HO, instead the red diamond highlights the statistical difference from GT. 
         }
         \label{Grafico_totale}
    \end{figure}

In the analysis of the results of the second part, in accordance with our expectations, the VFH condition was characterized by the highest level of weirdness compared to HO and GT conditions, which were in turn indistinguishable from one another  (Kruskal--Wallis test for all bootstrapping iterations: $p<10^{-17}$; post-hoc analysis: for HO-VFH $p<10^{-10}$ for all bootstrapping iterations, for GT-VFH $p<10^{-14}$ for all bootstrapping iterations, for GT-HO $p>0.05$ for 88 bootstrapping iterations out of 100, but the remaining has $p>0.01$). 
Figure~\ref{Grafico_totale}a illustrates the mean rank (in light of the bootstrapping procedure) in ``weirdness'' of motion (WM) along with its comparison interval~\cite{Hochberg1987}, in order to better highlight the significant difference among the three groups. 
Notably,  GT and HO  are indistinguishable from one another, while  VFH is significantly different from GT and HO.
Specifically, while VFH was considered ``weird'' in the majority of instances ($61$\%), GT was considered ``weird'' much less often than HO videos ($33$\% and $37$\%, respectively) (See Figure~\ref{part_2_analysis}).

We then asked the participants who detected weirdness in the videos to indicate which of the arrows exhibited such weirdness. We posit that more weirdness should be perceived in agents driven by algorithms than in agents associated with human beings. Our experiments indicated that the agent judged as weird was actually associated to a robot only in $16$\% of GT, while this proportion drastically increased to $47$\% of VFH (see Figure~\ref{part_2_analysis} patterned bars). This finding, combined with the Mann-Whitney test ($p<10^{-20}$ considering the whole bootstrapping analysis), supports the view that the trajectories generated by  GT  are perceived as much more natural than those generated by VFH. Additionally, it suggests that the motion of the robot controlled by  GT  is perceived as more human-like than the one generated by VFH. 

\begin{figure}[h]
    
         \centering
        \includegraphics[width=0.8\textwidth] 
         {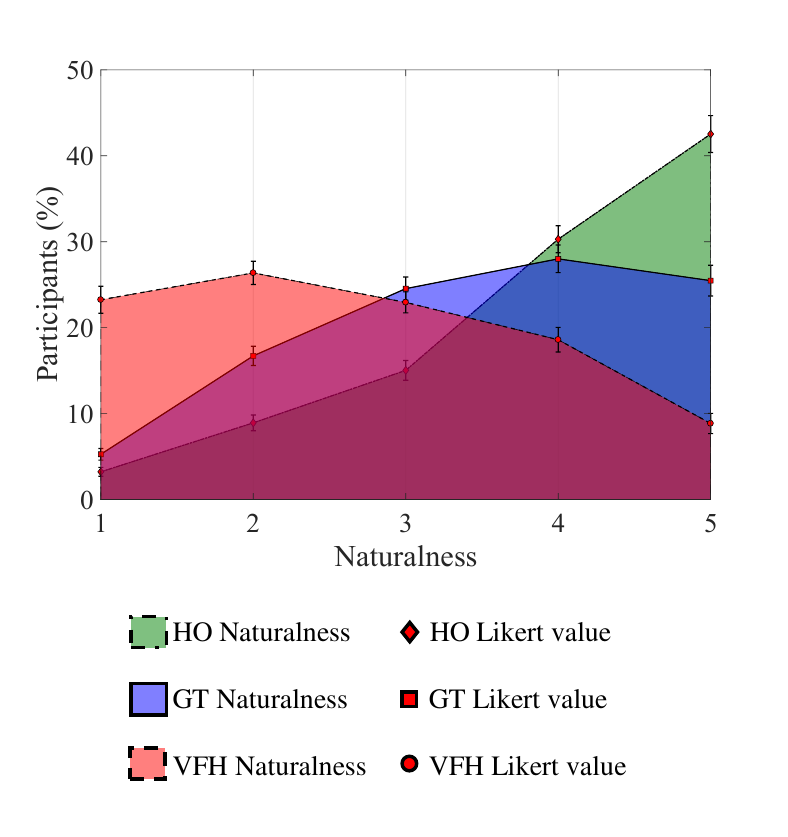}
         \caption{Result of the third part of the survey questionnaire.
         The participant assigns a degree of naturalness on a Likert scale from 1 (minimum naturalness) to 5 (maximum naturalness) considering the circled arrow in the 3 experimental conditions. The red points on the figure shows the average naturalness on each rate of the Likert scale considering 100 iterations with the bootstrapping approach, and the error bars represents the standard deviation. HO: the circled arrow is human; GT: the circled arrow is a robot driven by the game-theoretical trajectory planner; VFH: the circled arrow is driven by the Enhanced Vector Field Histogram algorithm.}
         \label{part_3_nat}
       
    \end{figure}

--In the third part, we further delved into the subjective rating of the three motion patterns by asking participants to focus on the motion of a circled target agent and evaluate whether such motion corresponds to a human or not (human likeness), along with its degree of naturalness on a Likert scale from one (minimum naturalness) to five (maximum naturalness). When focusing on the qualitative measurements of the human likeness, we observed that VFH-related arrows were considered much less human-like ($41.11$\%) than both GT ($64.59$\%) and HO ($80.31$\%). Thus, as illustrated in Figure~\ref{Grafico_totale}b, VFH is judged as the least human-like ($p<10^{-22}$ with Kruskal-Wallis bootstrapping; $p<10^{-5}$ VFH-GT, $p<10^{-30}$ VFH-HO post-hoc analysis), which is consistent with the previous part of the test, where VFH is perceived as generating the “weirdest” motion. Additionally, GT-related arrows were considered significantly less human-like compared to HO ($p<10^{-22}$ with Kruskal-Wallis bootstrapping; $p<10^{-4}$ post-hoc analysis GT-HO). We note that, in this case, the comparison intervals are not overlapping, thus denoting a clear distinguishability across experimental conditions. Figure~\ref{part_3_nat} illustrates the results related to the naturalness of the circled arrow. The figure shows the result about the average naturalness of motion of the circled arrow on a Likert scale from $1$ (minimum naturalness) to $5$ (maximum naturalness), computed over the 100 iterations of the bootstrapping procedure. According to our expectations, HO exhibits the highest mean degree of naturalness ($4$), closely followed by GT ($3.5$),  whereas a larger gap separates VFH ($2.6$). Importantly, a value of 3 in the adopted Likert scale was associated with the response “natural,” while values below this threshold were associated with a “non-natural” motion pattern. In light of this labeling, although HO was perceived as more ``natural'' than all other experimental conditions, it is noteworthy that GT was also classified as ``natural''. 

\section*{Discussion and conclusions}

The main goal of our study was to design a navigation system for autonomous robots moving through populated environments,  characterized by a high degree of acceptability by humans. Specifically, in light of the increasing use of autonomous robots in real life, we tested whether a navigation system designed through the principles of game theory would generate indistinguishable trajectories from those walked by human beings. To this aim, we first leveraged game theory to develop a model capable of predicting the intention of motion of humans in populated environments, and then, based on this model, we devised a trajectory planning algorithm for a mobile robot.  Finally, to assess the social acceptance of the generated robotic trajectories, we conducted a survey questionnaire on a statistically robust group of volunteers using a variation of the Turing test. 

For greater completeness and toward even more robust outcomes, before analyzing the results collected from the test, we also analyzed the geometrical features of the robotic trajectories, generated in the three experimental conditions (HO, GT, and VFH), selecting three metric parameters from the state of the art (PLR, PR, CPD). The ranking obtained through this analysis (HO, GT, VFH) is consistent with the results obtained through the Turing test, except for the closest pedestrian distance (CPD), in which the trajectories generated by our planner (GT) exhibit higher values of the parameter than those measured in environments populated by   humans only (HO). We hypothesize that this exception is due to the fact that our model guarantees by design a minimum safe distance to pedestrians to prevent collisions and to ensure, in any case, a comfortable action space. On the other hand, humans on the walk are more flexible in this respect, and evaluate circumstances on a case basis. While the outcome of the Turing test is consistent with the analysis of the metric  parameters of the trajectories, the statistical analysis (Kruskal-Wallis) executed  on the latter shows that this finding is not statistically significant. To explain this non-statistically significant result, we point out that the statistical analysis was conducted on only seven experiments per group, with differential degree of variability, and thus characterized by a limited statistical power. Moreover, in line with these considerations, we conducted a systematic analysis (Levene test) to evaluate the degree of variability of the different scenarios. In other terms, we evaluated the extent to which each algorithm generated videos that were similar to one another. This analysis revealed that there exists a significant variability with respect to the path regularity (PR), whereby the videos with the robot controlled by the VFH are the most variable, compared to the HO and GT experimental conditions. This finding suggests that the VFH algorithm is less predictable (i.e., it provides less regular results) than our algorithm and a real pedestrian.

The variant of the Turing test comprises a  first part that functions as a training phase. The second part comprises two consecutive phases. The first phase is devoted to compare the social acceptability of trajectories generated by either our game-theoretical algorithm (GT) or a state of the art algorithm (VFH) against a reference experimental condition, a complex social environment populated by humans only (HO). To this aim, participants were asked to say if they perceived weirdness in HO, GT, or VFH experimental conditions. The statistical test confirms that the perceived weirdness in trajectories in which only human subjects are involved is statistically indistinguishable from trajectories where the GT-controlled robot and human subjects coexist. Conversely, the videos in which the trajectories are generated by the VFH algorithm are perceived with a remarkably higher degree of overall weirdness compared with either HO or GT scenarios.

In the second phase of the second part of the test, participants were asked to indicate which is the perceived "weird" arrow, if any. In this regard, we observed that the  trajectories generated by the VFH algorithm were more frequently recognized as “weird” than those generated by our GT algorithm. 

In the third part of the test, participants were requested to focus on a circled arrow (a human in the HO experimental condition, a robot in GT and VFH ones), and were asked to evaluate whether or not the motion of the circled arrow corresponded to human recordings, and then rate their degree of naturalness. We observed that, while the arrow in VFH scenario was perceived as not human-like, the arrow controlled through GT was considered human-like, albeit not as human-like as the one rated in the HO experimental condition. We believe that this result is related to the fact that, in this part of the test, participants were asked to focus  on one arrow only, thus being biased toward detecting an artificial behavior. The same ranking between the three experimental conditions (HO, GT, and VFH) resulted from the analysis of the naturalness of motion of the circled arrow. Indeed, HO has the highest degree of naturalness, closely followed by our GT trajectory planner, and then by the VFH planner.  

We can conclude that, if participants are not guided to focus on a particular arrow, they would not distinguish much difference between a real human and a robot controlled through our game-theoretical framework and, therefore, the generated trajectory is a good candidate for social acceptance. This implies that our trajectory planning algorithm would help programming robots to blend well in populated environments, and, hence, to be perceived as more friendly, collaborative, and non-hostile.

Our findings are consistent with other studies in the literature, such as \cite{turnwald2019human}, where a different game-theoretical planner is perceived almost as human-like as human recordings. However, in \cite{turnwald2019human}, the authors created a human-like motion planner for mobile robots, still maintaining a simplified framework that does not comprise, for example, human groups, obstacle avoidance performed by humans, and the human desire of keeping a safe vital space around them \cite{hall1966hidden}. Moreover, their tests only comprise simplified scenarios: a first test with either only humans, or only robots; a second test in which the participant, based on virtual reality, interacts with an agent who can move as a human or a robot. In our study, we went one step further in  modeling (including the vital space, the group recognition, and the human-obstacle interaction) but also in the design of the variation of the Turing test (considering a real case scenario in which a robot moves in a human populated environment).
Nevertheless, it is hard to make extensive comparisons with other approaches, as the literature on variants of the Turing tests for assessing social acceptability of a robot agent is scant. 

Notably, the literature reports three main methods to evaluate the human-likeness and the social acceptance of robot navigation: (i) definition of social rules or performance metrics and, then, assessment of the adherence of the robot motion planner to these principles \cite{kirby2009companion,muller2008socially,pradeep2016human}; (ii) comparison between simulated trajectories and observed pedestrian behavior \cite{tamura2012development}; (iii) questionnaire based on a variation of the Turing test \cite{kretzschmar2016socially,turnwald2019human}. 
The main limitation of the first two methods is that they do not consider how humans perceive the robot. However, these methods can be applied to evaluate, as a preliminary test, some features of the generated trajectories. Indeed, our analysis of the metric parameters of the generated trajectories falls within the first methodology, whereas the second methodology has been used as a validation criterion for our game-theoretical model of pedestrian motion.

Hence, toward our aims, we deemed the Turing as an effective means to study the human-likeness and the social acceptability of the generated trajectories.
Unlike the Kretzschmar's \cite{kretzschmar2016socially} and Turnwald's \cite{turnwald2019human} tests, where volunteers watched  videos in which the totality of  agents moved either in an artificial way or  as real pedestrians, our questionnaire changes completely such a perspective. In fact, our test videos reproduce a true use case scenario of the algorithm (an environment populated by people with a single robot moving within), where the real nature of agents is masked and made uniform  to eliminate any participants' bias.
Moreover, unlike Kretzschmar's test \cite{kretzschmar2016socially}, where the Turing test is executed only on 10 participants, we performed an a priori power analysis to infer the correct sample size to obtain statistically significant results. Due to the largely superior size of collected data than the outcome of the power analysis, we carried out a 100-iteration bootstrap, always getting consistent results across iterations, highlighting the robustness of our results and further corroborating our hypothesis.\\

When interpreting the results of our study, we should also acknowledge the limitations of the \textbf{model} and of the \textbf {test} design. 
Regarding the former, our \textbf {model} does not take into account the uncertainties that arise from the interaction with the external world. Importantly, the stochasticity of human behavior is not explicitly modeled, although this is implicitly accounted for through tuning model parameters identified from real trajectory data, extracted from surveillance cameras. 
A range of simplifying assumptions were in order to handle the computational complexity of the algorithm. The main one resides in the discrete nature of our model, whereby each agent can choose between a fixed number of motion directions --an indispensable trade-off between predictive accuracy and computational effort. 
Moreover, the designed human motion model has been devised to operate with a limited number of pedestrians: its computational complexity may be difficult to manage if the number of agents increases to more than a dozen. 
The pedestrian model used in this study only considers people’s goal-directed and collision-avoiding behaviors, while ignoring other social activities that humans may perform in a pedestrian urban scenario, such as waiting for a bus or wandering without a clear direction. Thus, any pedestrian behavior that is not contemplated by our model breaks the assumptions under which our system works.
In addition, our method does not allow customization of trajectories. For example, the prediction of a trajectory walked by an  elderly person may be coincident with that of a child.

The main limitation of the \textbf{test} design is the choice of the navigation algorithm chosen for comparison (VFH). Ideally, more than an algorithm should have been selected in order to mitigate algorithm-induced biases. However, since the execution of the Turing test already took about 20 minutes to the average participant, we prefer to limit our comparison to only one algorithm at the state of the art, in order to avoid increasing the time of the experiment for each participant, mitigate attention biases and, in the end, achieve robust results.

Our work can be extend along several directions. To manage and predict the motion of big crowds, mean-field games could be adopted \cite{dogbe2010modeling}. We remark, however, that crowded and populated scenarios are different under many aspects, and the deployment of a robot in the two scenarios would cover totally different application fields. 

The lack of customization in the inference of trajectories by our model can be mitigated by combining our approach with learning strategies as in \cite{ma2017forecasting}, encompassing variegate behaviors across the experimental scenario.
In fact, adding variability to the pedestrian model might allow for a more accurate prediction of human motion pattern and should allow the robot to better adapt to the needs of the human with whom it is interacting.
For example, if a robot recognizes a person who has difficulties in walking, the robot should be able to predict their movement and possibly reduce its speed.
Moreover, it would be interesting to understand and assess the quality of our generated trajectories considering not only social acceptability but also the comfort \cite{shiomi2014towards} feeling of participants, for instance by creating a real shared real environment with humans and a robot.

\section*{Acknowledgments}
A.R. and G.G are  partially supported by Compagnia di San Paolo. 
S. G. is partially supported by the ERC under project COSMOS (802348).

The experimental protocol regulating the administration of the Turing test to human subjects, the evaluation of the results, and the data management plan was approved by the ethical committee of the \emph{Istituto Superiore di Sanit\`a} (Italian Institute of Health) with approval code \emph{AOO-ISS 10/07/2020 - 0024079, Class: PRE BIO CE 01.00.} 

Each participant also provided informed consent, after the explanation of the nature and possible consequences of the study.\\

The code used for the generation of the trajectories and the anonymized data collected during the experiments are available through the following link:

https://gitlab.com/PoliToComplexSystemLab/game-theoretic-trajectory-planning.git

\section*{Attributions}
Conceptualization SG SM AR

Data Curation GG SP

Formal Analysis GG SP

Funding Acquisition AR

Investigation All

Methodology SG SM AR 

Project Administration AR 

Resources SG SM AR

Software GG SP

Supervision SG SM AR

Validation GG SM

Visualization GG SP

Writing original draft GG AR

Writing review and editing All

\bibliographystyle{Science}

\bibliography{science}

\begin{thebibliography}{10}

\bibitem{torras2016service}
C.~Torras, {\it European Review\/} {\bf 24}, 17 (2016).

\bibitem{kruse2013human}
T.~Kruse, A.~K. Pandey, R.~Alami, A.~Kirsch, {\it Robotics and Autonomous
  Systems\/} {\bf 61}, 1726 (2013).

\bibitem{fox1997dynamic}
D.~Fox, W.~Burgard, S.~Thrun, {\it IEEE Robotics \& Automation Magazine\/} {\bf
  4}, 23 (1997).

\bibitem{fiorini1998motion}
P.~Fiorini, Z.~Shiller, {\it The International Journal of Robotics Research\/}
  {\bf 17}, 760 (1998).

\bibitem{van2008reciprocal}
J.~Van~den Berg, M.~Lin, D.~Manocha, {\it 2008 IEEE International Conference on
  Robotics and Automation\/} (IEEE, 2008), pp. 1928--1935.

\bibitem{kivrak2020social}
H.~Kivrak, F.~Cakmak, H.~Kose, S.~Yavuz, {\it Engineering Science and
  Technology, an International Journal\/}  (2020).

\bibitem{trautman2015robot}
P.~Trautman, J.~Ma, R.~M. Murray, A.~Krause, {\it The International Journal of
  Robotics Research\/} {\bf 34}, 335 (2015).

\bibitem{rios2015proxemics}
J.~Rios-Martinez, A.~Spalanzani, C.~Laugier, {\it International Journal of
  Social Robotics\/} {\bf 7}, 137 (2015).

\bibitem{turnwald2019human}
A.~Turnwald, Human-like motion planning in populated environments, Ph.D.
  thesis, Technische Universit{\"a}t M{\"u}nchen (2019).

\bibitem{sisbot2007human}
E.~A. Sisbot, L.~F. Marin-Urias, R.~Alami, T.~Simeon, {\it IEEE Transactions on
  Robotics\/} {\bf 23}, 874 (2007).

\bibitem{shiomi2014towards}
M.~Shiomi, F.~Zanlungo, K.~Hayashi, T.~Kanda, {\it International Journal of
  Social Robotics\/} {\bf 6}, 443 (2014).

\bibitem{chen2017socially}
Y.~F. Chen, M.~Everett, M.~Liu, J.~P. How, {\it 2017 IEEE/RSJ International
  Conference on Intelligent Robots and Systems (IROS)\/} (IEEE, 2017), pp.
  1343--1350.

\bibitem{helbing1995social}
D.~Helbing, P.~Molnar, {\it Physical review E\/} {\bf 51}, 4282 (1995).

\bibitem{tadokoro1995motion}
S.~Tadokoro, M.~Hayashi, Y.~Manabe, Y.~Nakami, T.~Takamori, {\it Proceedings
  1995 IEEE/RSJ International Conference on Intelligent Robots and Systems.
  Human Robot Interaction and Cooperative Robots\/} (IEEE, 1995), vol.~2, pp.
  518--523.

\bibitem{hoeller2007accompanying}
F.~Hoeller, D.~Schulz, M.~Moors, F.~E. Schneider, {\it 2007 IEEE/RSJ
  International Conference on Intelligent Robots and Systems\/} (IEEE, 2007),
  pp. 1260--1265.

\bibitem{bennewitz2005learning}
M.~Bennewitz, W.~Burgard, G.~Cielniak, S.~Thrun, {\it The International Journal
  of Robotics Research\/} {\bf 24}, 31 (2005).

\bibitem{alahi2016social}
A.~Alahi, {\it et~al.\/}, {\it Proceedings of the IEEE conference on computer
  vision and pattern recognition\/} (2016), pp. 961--971.

\bibitem{gupta2018social}
A.~Gupta, J.~Johnson, L.~Fei-Fei, S.~Savarese, A.~Alahi, {\it Proceedings of
  the IEEE Conference on Computer Vision and Pattern Recognition\/} (2018), pp.
  2255--2264.

\bibitem{liang2019peeking}
J.~Liang, L.~Jiang, J.~C. Niebles, A.~G. Hauptmann, L.~Fei-Fei, {\it
  Proceedings of the IEEE Conference on Computer Vision and Pattern
  Recognition\/} (2019), pp. 5725--5734.

\bibitem{turnwald2016understanding}
A.~Turnwald, D.~Althoff, D.~Wollherr, M.~Buss, {\it International Journal of
  Social Robotics\/} {\bf 8}, 331 (2016).

\bibitem{zhang1998motion}
H.~Zhang, V.~Kumar, J.~Ostrowski, {\it Proceedings. 1998 IEEE International
  Conference on Robotics and Automation (Cat. No. 98CH36146)\/} (IEEE, 1998),
  vol.~1, pp. 638--643.

\bibitem{gabler2017game}
V.~Gabler, T.~Stahl, G.~Huber, O.~Oguz, D.~Wollherr, {\it 2017 IEEE
  International Conference on Robotics and Automation (ICRA)\/} (IEEE, 2017),
  pp. 2897--2903.

\bibitem{dragan2017robot}
A.~D. Dragan, {\it arXiv preprint arXiv:1705.04226\/}  (2017).

\bibitem{nikolaidis2017mathematical}
S.~Nikolaidis, J.~Forlizzi, D.~Hsu, J.~Shah, S.~Srinivasa, {\it arXiv preprint
  arXiv:1707.02586\/}  (2017).

\bibitem{nash1951non}
J.~Nash, {\it Annals of mathematics\/} pp. 286--295 (1951).

\bibitem{epley2007seeing}
N.~Epley, A.~Waytz, J.~T. Cacioppo, {\it Psychological review\/} {\bf 114}, 864
  (2007).

\bibitem{roesler2021meta}
E.~Roesler, D.~Manzey, L.~Onnasch, {\it Science Robotics\/} {\bf 6}, eabj5425
  (2021).

\bibitem{waytz2010sees}
A.~Waytz, J.~Cacioppo, N.~Epley, {\it Perspectives on Psychological Science\/}
  {\bf 5}, 219 (2010).

\bibitem{hall1966hidden}
E.~T. Hall, {\it The hidden dimension\/}, vol. 609 (Garden City, NY: Doubleday,
  1966).

\bibitem{mavrogiannis2021core}
C.~Mavrogiannis, {\it et~al.\/}, {\it arXiv preprint arXiv:2103.05668\/}
  (2021).

\bibitem{xie2017learning}
D.~Xie, T.~Shu, S.~Todorovic, S.-C. Zhu, {\it IEEE transactions on pattern
  analysis and machine intelligence\/} {\bf 40}, 1639 (2017).

\bibitem{manual1985special}
H.~C. Manual, {\it Transportation Research Board, Washington, DC\/} {\bf 1},
  985 (1985).

\bibitem{ulrich1998vfh+}
I.~Ulrich, J.~Borenstein, {\it Proceedings. 1998 IEEE international conference
  on robotics and automation (Cat. No. 98CH36146)\/} (IEEE, 1998), vol.~2, pp.
  1572--1577.

\bibitem{bitgood2006not}
S.~Bitgood, S.~Dukes, {\it Environment and behavior\/} {\bf 38}, 394 (2006).

\bibitem{mcneill2002energetics}
R.~McNeill~Alexander, {\it American journal of human biology\/} {\bf 14}, 641
  (2002).

\bibitem{osborne1994course}
M.~J. Osborne, A.~Rubinstein, {\it A course in game theory\/} (MIT press,
  1994).

\bibitem{sagratella2017algorithms}
S.~Sagratella, {\it Computational Optimization and Applications\/} {\bf 68},
  689 (2017).

\bibitem{lerner2007crowds}
A.~Lerner, Y.~Chrysanthou, D.~Lischinski, {\it Computer graphics forum\/}
  (Wiley Online Library, 2007), vol.~26, pp. 655--664.

\bibitem{pellegrini2009you}
S.~Pellegrini, A.~Ess, K.~Schindler, L.~Van~Gool, {\it 2009 IEEE 12th
  International Conference on Computer Vision\/} (IEEE, 2009), pp. 261--268.

\bibitem{saygin2000turing}
A.~P. Saygin, I.~Cicekli, V.~Akman, {\it Minds and machines\/} {\bf 10}, 463
  (2000).

\bibitem{likert1932technique}
R.~Likert, {\it Archives of psychology\/}  (1932).

\bibitem{prajapati2010sample}
B.~Prajapati, M.~Dunne, R.~Armstrong, {\it Optometry today\/} {\bf 16}, 10
  (2010).

\bibitem{erdfelder1996gpower}
E.~Erdfelder, F.~Faul, A.~Buchner, {\it Behavior research methods, instruments,
  \& computers\/} {\bf 28}, 1 (1996).

\bibitem{corder2014nonparametric}
G.~W. Corder, D.~I. Foreman, {\it Nonparametric statistics: A step-by-step
  approach\/} (John Wiley \& Sons, 2014).

\bibitem{roberts2014student}
M.~Roberts, R.~Russo, {\it A student's guide to analysis of variance\/}
  (Routledge, 2014).

\bibitem{biswas2021socnavbench}
A.~Biswas, A.~Wang, G.~Silvera, A.~Steinfeld, H.~Admoni, {\it arXiv preprint
  arXiv:2103.00047\/}  (2021).

\bibitem{ostertagova2014methodology}
E.~Ostertagova, O.~Ostertag, J.~Kov{\'a}{\v{c}}, {\it Applied Mechanics and
  Materials\/} (Trans Tech Publ, 2014), vol. 611, pp. 115--120.

\bibitem{gastwirth2009impact}
J.~L. Gastwirth, Y.~R. Gel, W.~Miao, {\it Statistical Science\/} {\bf 24}, 343
  (2009).

\bibitem{efron1994introduction}
B.~Efron, R.~J. Tibshirani, {\it An introduction to the bootstrap\/} (CRC
  press, 1994).

\bibitem{johnson1999insignificance}
D.~H. Johnson, {\it The journal of wildlife management\/} pp. 763--772 (1999).

\bibitem{kraska2013nonparametric}
M.~Kraska-Miller, {\it Nonparametric statistics for social and behavioral
  sciences\/} (CRC Press, 2013).

\bibitem{foster2018introduction}
G.~C. Foster, {\it et~al.\/}  (2018).

\bibitem{mann1947test}
H.~B. Mann, D.~R. Whitney, {\it The annals of mathematical statistics\/} pp.
  50--60 (1947).

\bibitem{fay2010wilcoxon}
M.~P. Fay, M.~A. Proschan, {\it Statistics surveys\/} {\bf 4}, 1 (2010).

\bibitem{Hochberg1987}
Y.~Hochberg, A.~Tamhane, {\it Multiple Comparison Procedures\/} (John Wiley,
  1987).

\bibitem{kirby2009companion}
R.~Kirby, R.~Simmons, J.~Forlizzi, {\it RO-MAN 2009-The 18th IEEE International
  Symposium on Robot and Human Interactive Communication\/} (IEEE, 2009), pp.
  607--612.

\bibitem{muller2008socially}
J.~M{\"u}ller, C.~Stachniss, K.~O. Arras, W.~Burgard, {\it Proc. of
  International Conference on Cognitive Systems\/} (2008).

\bibitem{pradeep2016human}
Y.~C. Pradeep, Z.~Ming, M.~Del~Rosario, P.~C. Chen, {\it 2016 IEEE
  International Conference on Mechatronics and Automation\/} (IEEE, 2016), pp.
  971--976.

\bibitem{tamura2012development}
Y.~Tamura, {\it et~al.\/}, {\it 2012 IEEE/RSJ International Conference on
  Intelligent Robots and Systems\/} (IEEE, 2012), pp. 382--387.

\bibitem{kretzschmar2016socially}
H.~Kretzschmar, M.~Spies, C.~Sprunk, W.~Burgard, {\it The International Journal
  of Robotics Research\/} {\bf 35}, 1289 (2016).

\bibitem{dogbe2010modeling}
C.~Dogb{\'e}, {\it Mathematical and Computer Modelling\/} {\bf 52}, 1506
  (2010).

\bibitem{ma2017forecasting}
W.-C. Ma, D.-A. Huang, N.~Lee, K.~M. Kitani, {\it Proceedings of the IEEE
  Conference on Computer Vision and Pattern Recognition\/} (2017), pp.
  774--782.

\end{thebibliography}

\captionsetup[figure]{labelfont={bf}}

\clearpage

\end{document}